\documentclass[aps,onecolumn]{revtex4-1}

\usepackage{hyperref}
\usepackage{graphicx}
\usepackage{amsmath,amssymb,amsfonts}

\begin{document}

\title{Exploiting individual differences to bootstrap communication}
\date{24th July 2025}

\author{Richard A. Blythe}
\email{r.a.blythe@ed.ac.uk}
\affiliation{School of Physics and Astronomy, University of Edinburgh, Edinburgh, UK}
\affiliation{Centre for Language Evolution, University of Edinburgh, Edinburgh, UK}

\author{Casimir Fisch}
\email{casimir.fisch@env.ethz.ch}
\affiliation{School of Physics and Astronomy, University of Edinburgh, Edinburgh, UK}
\affiliation{Institute for Atmospheric and Climate Sciences, ETH Z\"urich, Z\"urich, Switzerland}

\begin{abstract}
Establishing a communication system is hard because the intended meaning of a signal is unknown to its receiver when first produced, and the signaller also has no idea how that signal will be interpreted. Most theories for the emergence of communication rely on feedback to reinforce behaviours that have led to successful communication in the past. However, providing such feedback requires already being able to communicate the meaning that was intended or interpreted. Therefore these accounts cannot explain how communication can be bootstrapped from non-communicative behaviours. Here we present a model that shows how a communication system, capable of expressing an unbounded number of meanings, can emerge as a result of individual behavioural differences in a large population without any pre-existing means to determine communicative success or strong prior constraints on language structure. The two key cognitive capabilities responsible for this outcome are learning to behave predictably in a given situation, and an alignment of psychological states ahead of signal production that derives from shared intentionality. Since both capabilities can exist independently of communication, our results are compatible with theories in which large flexible socially-learned communication systems like language are the product of a general but well-developed capacity for social cognition.
\end{abstract}

\maketitle

\section{Introduction}

Human language is a large and flexible communication system that relates signals to meanings in a consistent manner across societies of up to millions of speakers \cite{Clark1996,Tomasello2008}. These socially-learned mappings are established by convention \cite{Clark1996,Lewis2002,Beckner2009,Hawkins2019}: in a communicative interaction, the signaller appeals to their experience to choose a signal that is likely to convey their desired intention, and the receiver does the same to draw a plausible interpretation. A fundamental question, relevant both to the acquisition \cite{Bloom2000,Tomasello2005,Chater2018} and origins \cite{Hurford1989,Tomasello2008,Fitch2010,Hurford2014} of language, is how agents can agree on which of a large repertoire of signals maps onto one of a potentially infinite number of meanings \cite{Quine1960,Blythe2016} when no convention has previously been established. 

Many models of convention formation through social learning rely on feedback, such as pointing, to confirm whether a signaller's intended meaning was correctly interpreted by the receiver \cite{Steels2003,Puglisi2008,Sukhbaatar2016,Spike2017}. In evolutionary game theoretic terms \cite{Nowak2006}, feedback delivers a reward for success that allows lucky guesses between a pair of agents to be amplified into community-wide conventions. However, such accounts fall foul of a \emph{signal redundancy paradox} \cite{Smith2005}: the ability to provide this feedback presupposes that a convention for the intended or interpreted meaning already exists. For example, if someone rubbing their stomach is sufficient to inform a receiver that the signaller is in an otherwise unobservable state of hunger, using the word `hungry' redundantly duplicates communication that is already possible.

Analyses of innately-specified animal call systems, such as that employed by vervet monkeys to evade different predator types \cite{Seyfarth1980}, sidestep this difficulty by eshewing any commitment to internalised meanings \cite{Lewis2002}. Instead, these analyses are couched in terms of actions and reactions performed by signaller and receiver, respectively \cite{Donaldson2007,ScottPhillips2012}. Nevertheless, predation is a response to pre-existing observable behaviour and communicates a failure to react appropriately to the signaller's action in a manner that is at least as forceful as pointing. Emergence of communication through natural selection is therefore afflicted by the same signal redundancy paradox. The fundamental question we address in this work is if there is any way to bootstrap communicative conventions without some form of communication already being in place. 

The proposal that we pursue here is statistical learning of signal and meaning co-occurrences \cite{Siskind1996,Isbilen2022,Roembke2023}. By reusing a signal that has, by chance, been produced more often in a given context, a convention might emerge blind of any overt agreement on its meaning. Returning to the earlier example, if several hours have passed since signaller and receiver last ate, both are likely to feel hungry at the same time, and a receiver may correlate a signaller's stomach-rubbing gesture or the word `hungry' with their own unobservable sensations independently of any pre-existing means to communicate them. Repeating this gesture when experiencing such sensations in the future might then allow its association with an intentional behaviour to grow. Communication would then be said to exist when such associations are shared across the society. 

Although this type of statistical learning has proven powerful for children to learn a set of pre-existing conventions from adults \cite{Blythe2010,Reisenauer2013,Blythe2016}, evidence of its ability to build a communication system from scratch is equivocal. On the one hand, iterated artificial language learning experiments have shown that human participants introduce systematicity into initially random mappings between words and objects, despite never receiving feedback about production errors \cite{Kirby2008}. Simulations replicate this finding in very small societies \cite{Smith2001,Kwisthout2008}. However shared systems fail to get off the ground in larger societies \cite{Vogt2003}. Even with just two agents, the system that emerges lacks structure: a signal is assigned randomly to each meaning with no regard to utilising the full signal repertoire \cite{Fontanari2011}. 

Taken together, these findings suggest that large societies of statistical learners might need to be equipped with strong expectations or constraints on language structure \cite{Bloom2000,Saffran2003} to build rich and effective communication systems \cite{DeBeule2006}. Examples of such constraints include an expectation that a word is more likely to describe a whole object rather than one of its parts \cite{Macnamara1972}, or that a novel term is more likely to describe a novel object \cite{Markman1988}. An alternative is that various social cognitive mechanisms of \emph{shared intentionality} \cite{Tomasello2007} are sufficient to constrain the candidate meanings for a signal. Prominent among these is \emph{joint attention} \cite{Tomasello1986}, whereby interacting agents are aware that they share a common focus of attention, which then provides candidate meanings for any signal that is produced. Neuroimaging studies  \cite{Stolk2016,Schilbach2025} suggest that such correlated mental states can be created ahead of signal production, indicating that if communication emerges through such interactions, it can be bootstrapped from non-communicative capabilities, rather than relying on a pre-existing channel for communicating success or failure.

In pursuing this hypothesis, we further determine how tightly attention needs to be constrained to a small range of candidate meanings  for communication to emerge, and how much the attention of interacting agents needs to overlap. To this end, we construct a model, described in Section~\ref{sec:meth} below, that allows for agents' attention to different possible meanings to be carefully controlled. To aid interpretability of the results, we further ensure that the mechanism for learning any associated signalling behaviour is grounded in the solid principles of Bayesian inference \cite{Griffiths2010,Gill2015} and information theory \cite{Cover2006}, as opposed to inventing ad-hoc rules. The final key component of this model is that agents behave predictably in a given situation \cite{Tomasello2005,Bar2011,Pickering2013,Chater2018,ContrerasKallens2024}.

By applying a mathematical analysis, we compare how shared communication emerges in three different scenarios, and find that a qualitatively different cultural evolutionary mechanism operates in each case. When joint attention is tightly constrained to a single meaning, we find that signals become conventionalised by neutral evolutionary forces analogous to genetic drift \cite{Kimura1985}. This explains the lack of structure previously identified in this regime \cite{Fontanari2011}. When attention is entirely unconstrained, and feedback is used to resolve the resulting ambiguity, signals that are most likely to be correctly interpreted are directly selected for. On the other hand, when candidate meanings are weakly constrained, a shared communication system can emerge as long as interacting agents distribute their attention over these candidate meanings in a sufficiently similar way. However, the selection pressure here is indirect, and arises only if there is sufficient variability in individual behaviour across the population. We further find that this mechanism for bootstrapping communication from non-communicative social cognitive capabilities scales to arbitrarily many meanings and signals, no matter how large the society, contrary to what was previously suggested by simulations \cite{Vogt2003}.

\section{Methods}
\label{sec:meth}

We begin by setting out the principles behind the model that we introduce in this work, along with the details as to how they are implemented.

\subsection{Distribution of attention}

Intentional signalling behaviour evolves over time through repeated interactions between a pair of agents drawn from a society of $N$ agents. In the simulation results shown below, these pairs are drawn uniformly at random (the interaction network is a complete graph), but the analytical results hold for a much wider class of interaction networks (see Appendix).

\begin{figure}[h!]
\begin{center}\includegraphics[width=0.75\linewidth]{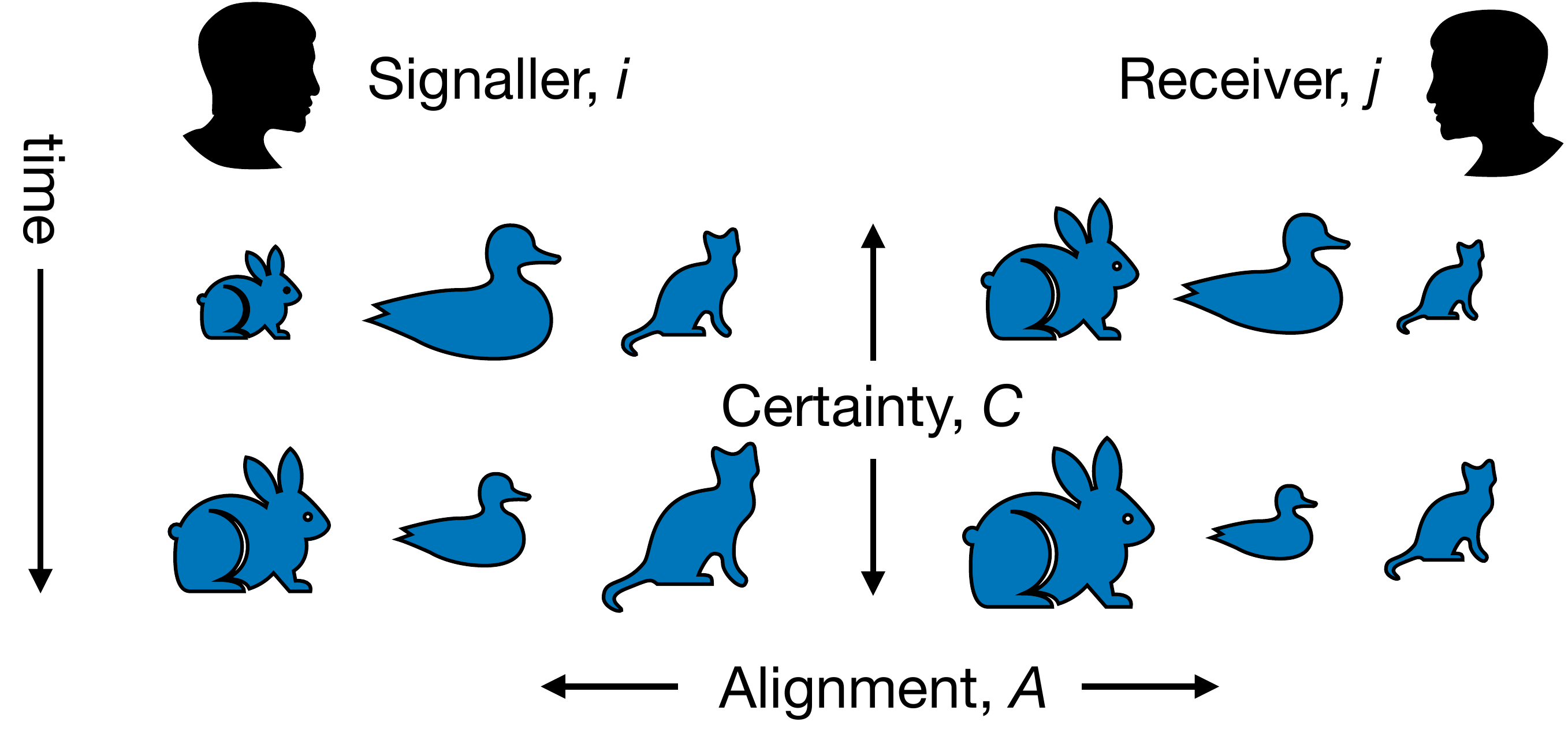}\end{center}
\caption{{\bf Attention of signaller and receiver to specific meanings}
In each interaction, signaller $i$ and receiver $j$ distribute their attention over a set of meanings (shapes) with varying weights (indicated by size). Variation between interactions at different times for the same agent is quantified by the certainty $0\le C\le1$; variation between signaller and receiver at the same time by the alignment $0\le A\le1$.}
\label{fig:attn}
\end{figure}

In an interaction, one agent is designated the signaller (and labelled $i$) and the other the receiver ($j$). Both have their attention distributed over a set of $M$ possible meanings. We denote the amount of attention given by the signaller to meaning $m$ as $\rho_i(m)$ and the amount given by the receiver as $\rho_j(m)$. See Fig~\ref{fig:attn}. Note that the attention to a given meaning varies between interactions: we have, however, suppressed an explicit dependence on time to keep the notation manageable. These attentional weights are normalised so that they always sum to unity for each agent in every interaction: $\sum_{m} \rho_{\ell}(m)=1$ for all $\ell$. 

There are two important statistical properties of the distribution over attention. The first of these, the \emph{certainty} $C$, quantifies how much the attention to specific meanings varies between interactions. For example, in the earlier interaction shown in Fig~\ref{fig:attn}, the signaller is attending more strongly to the duck than in the later one. The weight $\rho_i(m)$ determines the probability that they select meaning $m$ as the \emph{topic} of the interaction, that is, the meaning that they wish to communicate. When attention varies strongly between interactions, it must necessarily also vary between meanings within each interaction, as the total amount of attention is fixed. This means we can also interpret the certainty as a measure of how strongly constrained an agents' beliefs are about the meaning of a signal produced in an interaction. For example, where strong constraints (such as whole-object or mutual exclusivity constraints \cite{Macnamara1972,Markman1988}) are operating, attention is likely to be drawn to a small number of meanings that are specific to a given interaction.

In accordance with the above, we define the certainty as
\begin{equation}
\label{C}
C_\ell = \frac{\sum_m {\rm Var}[\rho_\ell(m)]}{1 - \sum_m \mathbb{E}[\rho_\ell(m)]^2} 
\end{equation}
where the expectation value and variance are taken over all possible interactions involving agent $\ell$. The numerator specifies how much more likely it is for two topics to be the same if they are sampled with replacement from the same set of attentional weights (i.e., at the same time) than if they are drawn from two independent sets of attentional weights (i.e., at different times). The denominator is the maximum value that the numerator can take (occurring when attention is always focussed on a single meaning). Thus $C_\ell$ is normalised to the range $0 \le C_\ell \le 1$, with $C_{\ell}=0$ corresponding to the situation where the attention distribution is static and completely unconstrained, and $C_\ell=1$ to where it is maximally variable and tightly constrained. We assume that all agents experience the same overall level of certainty, i.e.,$C_{\ell} = C$, a common value for all agents $\ell$.

The proposed mechanism for the emergence of communication without feedback is that the signaller's behaviour might correlate with the meanings that the \emph{receiver} is attending to. The second key property therefore is how similar the signaller's and receiver's attention distributions are. For example, in Fig~\ref{fig:attn}, although the weights are not exactly the same, we see that both agents attend more to the duck in the first interaction than in the second, while they attend more to the rabbit in the second than the first. Mechanisms of shared intentionality \cite{Tomasello2007}, including joint attention \cite{Tomasello1986,Kwisthout2008}, can generate this \emph{alignment} of the attentional weights.

We quantify the level of alignment between different agents $i$ and $j$ with the parameter
\begin{equation}
\label{A}
A_{ij} = \frac{\sum_m {\rm Cov}[ \rho_i(m), \rho_j(m) ]}{\sum_m \sqrt{ {\rm Var} [ \rho_i(m) ]\, {\rm Var} [ \rho_j(m) ]} } \;,
\end{equation}
which is normalised so that $-1 \le A_{ij} \le 1$. As with the certainty we assume that the alignment between any pair of agents takes the same value, $A_{ij}=A$ when $i\ne j$. Although not exactly the same as Pearson's correlation coefficient, its interpretation is similar. First, the numerator equals the denominator when both agents always direct their attention in exactly the same way over the set of meanings (i.e., $\rho_i(m)=\rho_j(m)$ always): this is the maximum alignment that can be achieved given the underlying level of certainty, $A=1$. On the other hand, if there is no systematic relationship between the signaller's and receiver's attentional weights, $A=0$. In this case, there is no way that the signaller's observable behaviour could vary systematically with the receiver's attention, and we expect it to be impossible for communication to emerge in the absence of communicative feedback in this case.

In the following, we assume that each agent attends to meaning $m$ with overall weight $\rho(m)$ when averaged over many interactions, that is, there is no long-term difference between agents in \emph{how} they distribute their attention, only \emph{when}. Our mathematical treatment allows this average weight to vary arbitrarily between meanings, although we restrict to the case of the uniform distribution $\rho(m)=\frac{1}{M}$ in simulations and when analysing the emergence of communication. To realise a specific combination of $C$ and $A$ in simulations (the latter restricted to $0\le A \le1$), we drew the set of weights for the signaller from a Dirichlet distribution so that the desired variance in (\ref{C}) was obtained. Then, with probability $A$ we assigned exactly the same set of weights to the receiver; otherwise (with probability $1-A$) we sampled an independent set of weights from the same Dirichlet distribution. Note however that our mathematical analysis does not rely on this specific construction, and in the three regimes we consider, applies to any distribution of weights with the same statistical properties.

\subsection{Topic selection and signal production}

\begin{figure}[h!]
\begin{center}\includegraphics[width=0.5\linewidth]{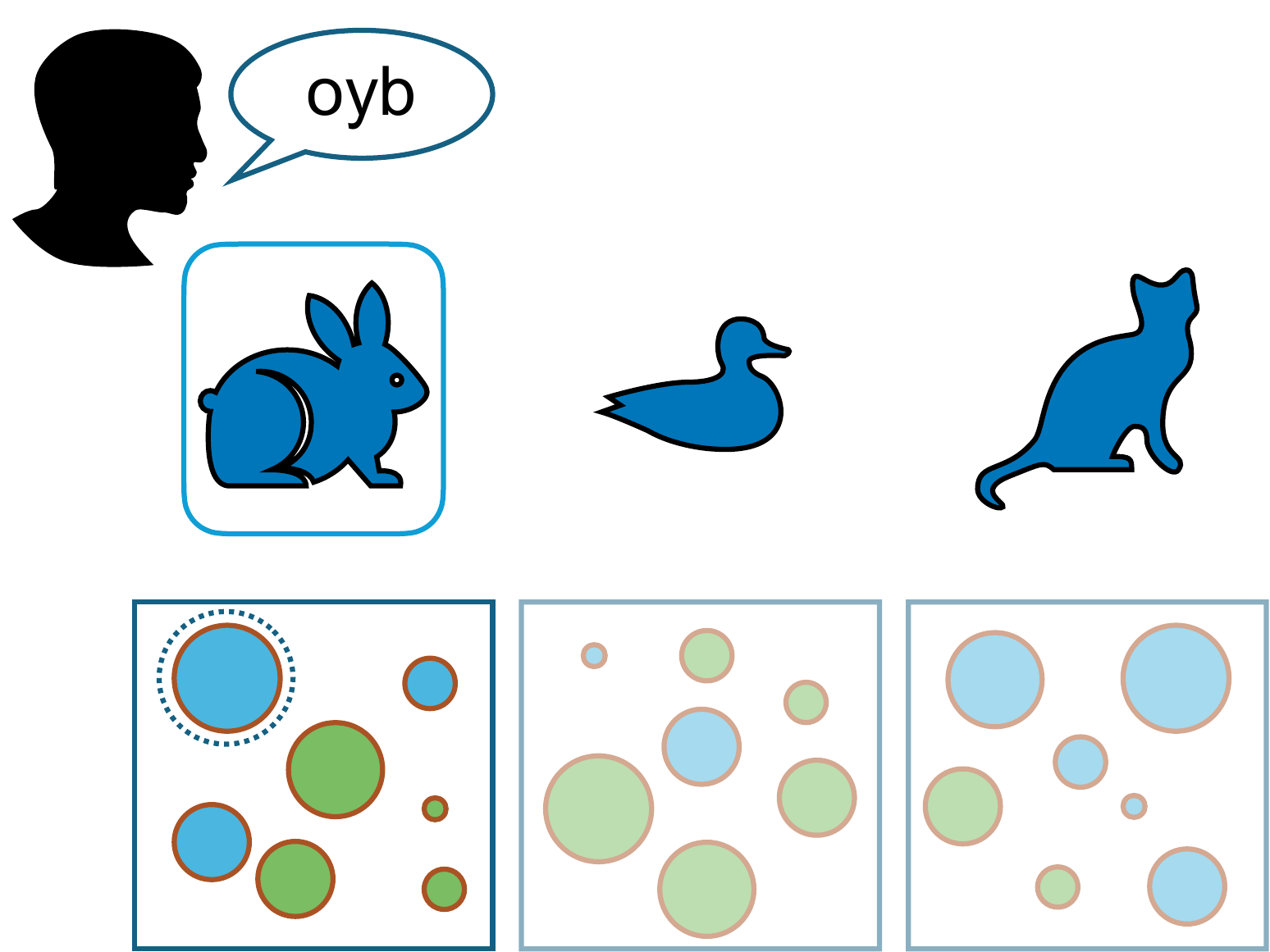}\end{center}
\caption{{\bf Topic selection and signal production}
The signaller samples a topic (here, the rabbit) in proportion to the attentional weight (size of the corresponding shape). The lower row of boxes indicates memory of which signal (different coloured circles) was previously interpreted by the signaller as having the corresponding meaning. Only the memories related to the topic are relevant (irrelevant memories are shaded). Memories decay over time, the size of each circle corresponding to the strength of the memory that remains. The signaller samples a signal in proportion to its strength in the memory. Here a blue circle is sampled, which corresponds to the verbalisation `oyb'.}
\label{fig:produce}
\end{figure}

As noted above, the signaller $i$ selects the topic of the interaction by sampling from their attentional weights $\rho_i(m)$. They are further equipped with a memory of previous interactions, partitioned according to the meanings that they experienced as receivers by employing Bayesian inference strategies set out below. This partitioning is illustrated in Fig~\ref{fig:produce} through the lower row of boxes. Each circle within a box corresponds to a memory of a signal previously interpreted as the corresponding meaning, with more recent memories being stronger (larger circles) than more distant ones. The size weighted distribution over signals for meaning $m$ is denoted $\phi_i(s|m)$, normalised so that $\sum_s \phi_i(s|m) = 1$. A central assumption in the model is that agents are motivated to behave predictably  \cite{Tomasello2005,Chater2018,ContrerasKallens2024}. Concretely, this means that agents conform to their own beliefs about each other's behaviour, and produce signal $s$ for meaning $m$ with probability $\phi_i(s|m)$. Note that although the signal is illustrated in Fig~\ref{fig:produce} as a verbalisation, it could be any other intentional behaviour, such as a facial expression or a hand gesture.

\subsection{Signal interpretation}

The task faced by the receiver is the inverse to that of the signaller: instead of choosing a signal to represent a meaning sampled from the attentional weights, they must infer the topic given the signal and their own attentional weights. Bayesian inference \cite{Griffiths2010,Gill2015} provides a natural mathematical framework for performing such probabilistic reasoning and is thus a widely adopted paradigm in cognitive science.

\begin{figure}[h!]
\begin{center}\includegraphics[width=0.5\linewidth]{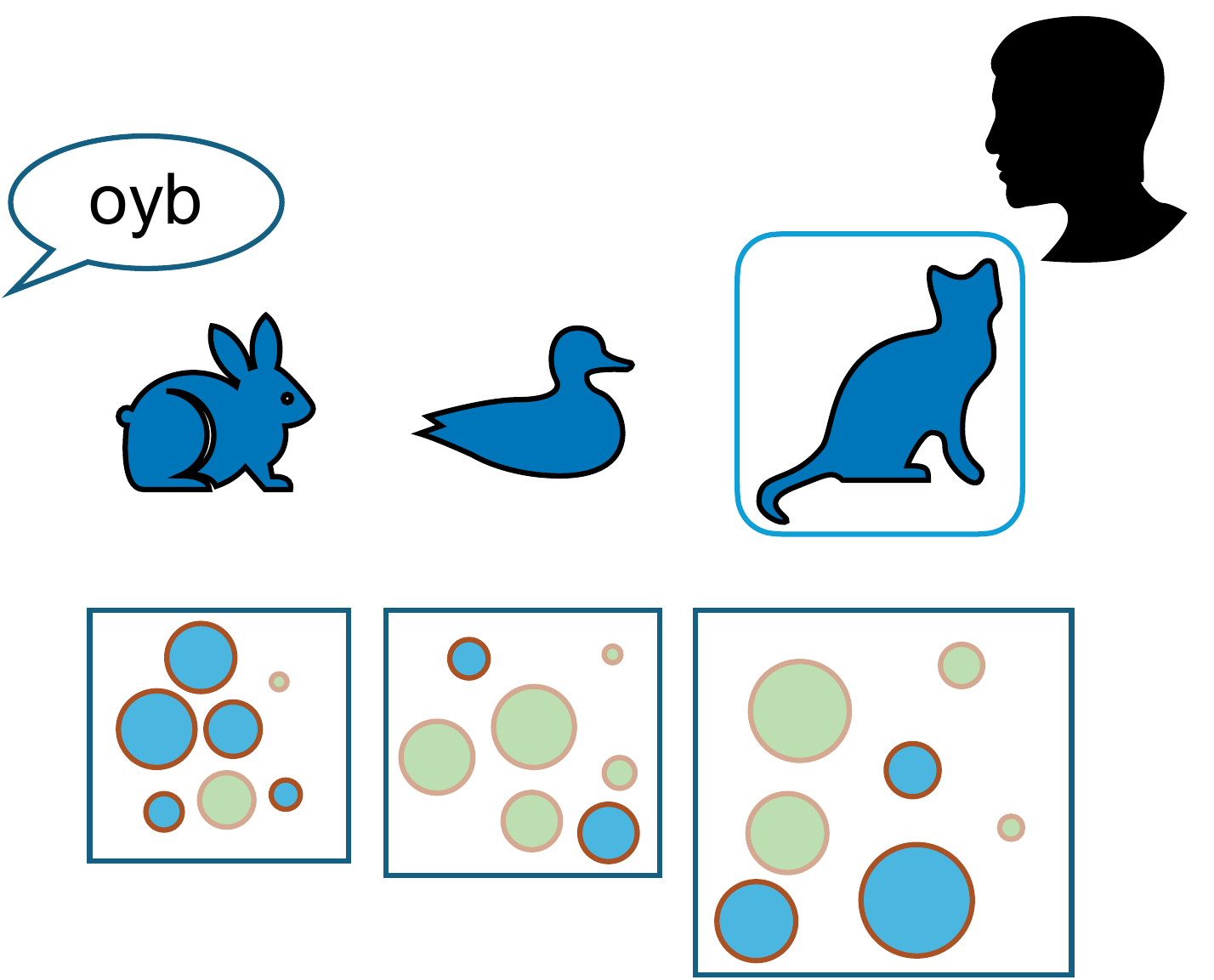}\end{center}
\caption{{\bf Signal interpretation} The receiver interprets the signal (here, the vocalisation `oyb') by focussing on all memories of that signal's production in the past (blue circles). Memories of other signals are irrelevant (hence shaded). These memories are combined with the receiver's attentional weights by scaling their size in proportion to those weights (shown as each box, and its contents, scaled accordingly). The interpreted meaning (here, cat) is sampled according to the resulting combined weight.}
\label{fig:interp}
\end{figure}

Here, it manifests as the receiver's attentional weights constituting a set of prior beliefs for the meaning of the signal they have encountered. When weighted by their estimate $\phi_j(s|m)$ of how likely it is that a member of their society would use signal $s$ to convey meaning $m$ (which, as noted above, corresponds also to the probability that they themselves would use this signal for that meaning), we obtain the posterior distribution over \emph{interpretations}
\begin{equation}
\label{psi}
\psi_j(m|s) = \frac{ \phi_j(s|m) \rho_j(m) }{ \sum_\mu \phi_j(s|\mu) \rho_j(\mu) } \;.
\end{equation}
See Fig~\ref{fig:interp}. This posterior distribution accounts in a principled way both for the receiver's expectations as to likely topics of conversation and their experience of signalling behaviour in their society.

\subsection{Predicting signalling behaviour with finite memory}

At the heart of any model for the cultural evolution of signalling behaviour is a process by which signalling behaviour is learned (see e.g.~\cite{Spike2017} for a survey). Although details vary considerably, the unifying premise is that when a signal is produced in the context of a given meaning, a memory of this is retained such that the same signal is more likely to be produced in the future when the agent seeks to communicate that same meaning. Here we follow the principled approach of \cite{Reali2010}, which is also grounded in Bayesian inference. The idea is that, given a sequence of $T$ signals that have been interpreted with meaning $m$, and a prior distribution over the $S$ available signals, the agent $\ell$ computes the \emph{posterior predictive distribution} \cite{Gill2015} that expresses how likely the next signal with meaning $m$ is $s$. This furnishes the distribution over signaller's productions, $\phi_{\ell}(s|m)$, introduced above. A limitation of the formalism of \cite{Reali2010} is that it is restricted to the case of \emph{batch} learning: that is, an agent acts only as a receiver while being exposed to the $T$ signals, after which they mature and act only as a signaller, producing signals from an unchanging distribution $\phi_{\ell}(s|m)$. In the present work, we require agents to act as signallers and receivers concurrently, engaging therefore in \emph{online} learning and with their signalling behaviour evolving over time. Moreover, we also seek to build in a finite-memory constraint, such that older interactions carry less weight than newer ones. This feature is important to avoid the system becoming trapped in suboptimal states \cite{Spike2017}.

The prior that is adopted by \cite{Reali2010} is a Dirichlet distribution over sets of frequency estimates $\{ \hat\phi_{\ell}(s|m) \}$ for each meaning $m$ and agent $\ell$. Here we use a hat to distinguish these conjectured frequencies from the posterior predictive distribution $\phi_\ell(s|m)$. We assume that all signals are initially considered interchangeable, which implies a prior of the form
\begin{equation}
\mathbb{P}_0(\{\hat\phi_{\ell}(s|m)\}) \propto \prod_s \hat\phi_{\ell}(s|m)^{\frac{\alpha}{S}-1} \;,
\end{equation}
where we have omitted the normalisation to avoid undue notational clutter. The parameter $\alpha$ determines how much variability an agent expects in signalling behaviour, with small values of $\alpha$ corresponding to an expectation that one signal is very much more likely to be used than any other for a meaning, while larger values of $\alpha$ allow for multiple signals to be used variably. The value of $\alpha$ thus quantifies the strength of a prior constraint on the structure of a nascent communication system.

For batch learning, the likelihood of a sequence of signals $(\sigma_1, \sigma_2, \ldots, \sigma_L)$ being produced is obtained by multiplying the conjectured frequency of each one:
\begin{equation}
\label{L}
\mathbb{L}(\sigma_1, \ldots, \sigma_L) = \hat\phi_{\ell}(\sigma_1|m) \hat\phi_{\ell}(\sigma_2|m) \cdots \hat\phi_{\ell}(\sigma_T|m) \;.
\end{equation}
The posterior predictive distribution $\phi_{\ell}(s|m)$ is then obtained by averaging the frequency $\hat\phi_{\ell}(s|m)$ over the posterior obtained by multiplying the prior and the likelihood (and normalising appropriately) \cite{Reali2010}.

To incorporate memory loss, we appeal to information theory, specifically, that in an optimal encoding of the conjectured signal distribution, a memory of signal $s$ consumes $-\log_2 \hat\phi_{\ell}(s|m)$ bits of memory \cite{Cover2006}. Suppose now that a random fraction $\lambda$ of the bits in the memory are deleted whenever new information is stored. Then, the string that was stored for a signal $s$ gets converted into a memory of a different signal with the shorter length $-(1-\lambda) \log_2 \hat\phi_{\ell}(s|m)$, and thus has a higher frequency. This corresponds to replacing each frequency $\hat\phi_{\ell}(\sigma_t|m)$ in the likelihood (\ref{L}) with the modified frequency $\hat\phi_{\ell}(\sigma_t|m)^{\epsilon_t}$ where $\epsilon_t=(1-\lambda)^{T-t}$.

Given these definitions, one finds (by using the properties of the Dirichlet distribution \cite{Reali2010}) that after $T$ interactions, the posterior predictive distribution is
\begin{equation}
\label{ppd}
\phi_{\ell}(s|m)= \frac{\sum_{t=1}^T \omega_{m\mu_t}\delta_{s\sigma_t} (1-\lambda)^{T-t}  + \frac{\alpha}{S}}{\sum_{t=1}^{T} \omega_{m\mu_t} (1-\lambda)^{T-t}+\alpha} \;.
\end{equation}
Here, $\delta_{s\sigma_t}=1$ if the signal produced in interaction $t$, $\sigma_t$, is $s$, otherwise it is zero. Meanwhile $\omega_{m\mu_t}=1$ if the agent chose to store information about the interaction at time $t$, otherwise it is zero. By default, a receiver stores information after every interaction. If desired, we can also incorporate communicative feedback into this process by storing new information \emph{only} when communication is a success: in such a case, we take $\omega_{m\mu_t}=1$ only if the interpretation $m$ matches the topic $\mu_t$. If it is a failure, memory of previous signals is assumed to decay through the same mechanism as above, but nothing is stored to replace it. The numerator of (\ref{ppd}) can be recognised as set of counts of interactions where $\mu_t$ was the topic, $\sigma_t$ was the signal produced, and $m$ was the meaning subsequently inferred, these counts weighted exponentially by the time that has passed since that interaction. The denominator ensures normalisation. The case considered by \cite{Reali2010} corresponds to no memory decay, $\lambda=0$, and all the weights in the sum are equal to one.

It is more helpful for both simulation and mathematical analysis to work with the \emph{change} in the posterior predictive distribution that occurs as a result of the interaction that takes place at time $T$. By subtracting  (\ref{ppd}) at two successive time points, we find
\begin{equation}
\label{update}
\delta\phi_{\ell}(s|m) = \frac{\omega_{m\mu_T} [\delta_{s\sigma_T}-\phi_{\ell}(s|m)] + \lambda\alpha[\frac{1}{S}-\phi_{\ell}(s|m)]}{\sum_{t=1}^{T} \omega_{m\mu_t} (1-\lambda)^{T-t} + \alpha} \;.
\end{equation}
When the rate of information loss is small, the sum in the denominator converges to $\frac{\beta_{\ell}(m)}{\lambda}$ as $T\to\infty$, where $\beta_{\ell}(m)$ is the probability that $\omega=1$ conditioned on $m$ being interpreted. When there is no communicative feedback, $\beta_{\ell}(m)=1$ always, whereas when feedback is available, its value depends on the state of the system and changes over time.

\begin{figure}[h!]
\begin{center}\includegraphics[width=0.5\linewidth]{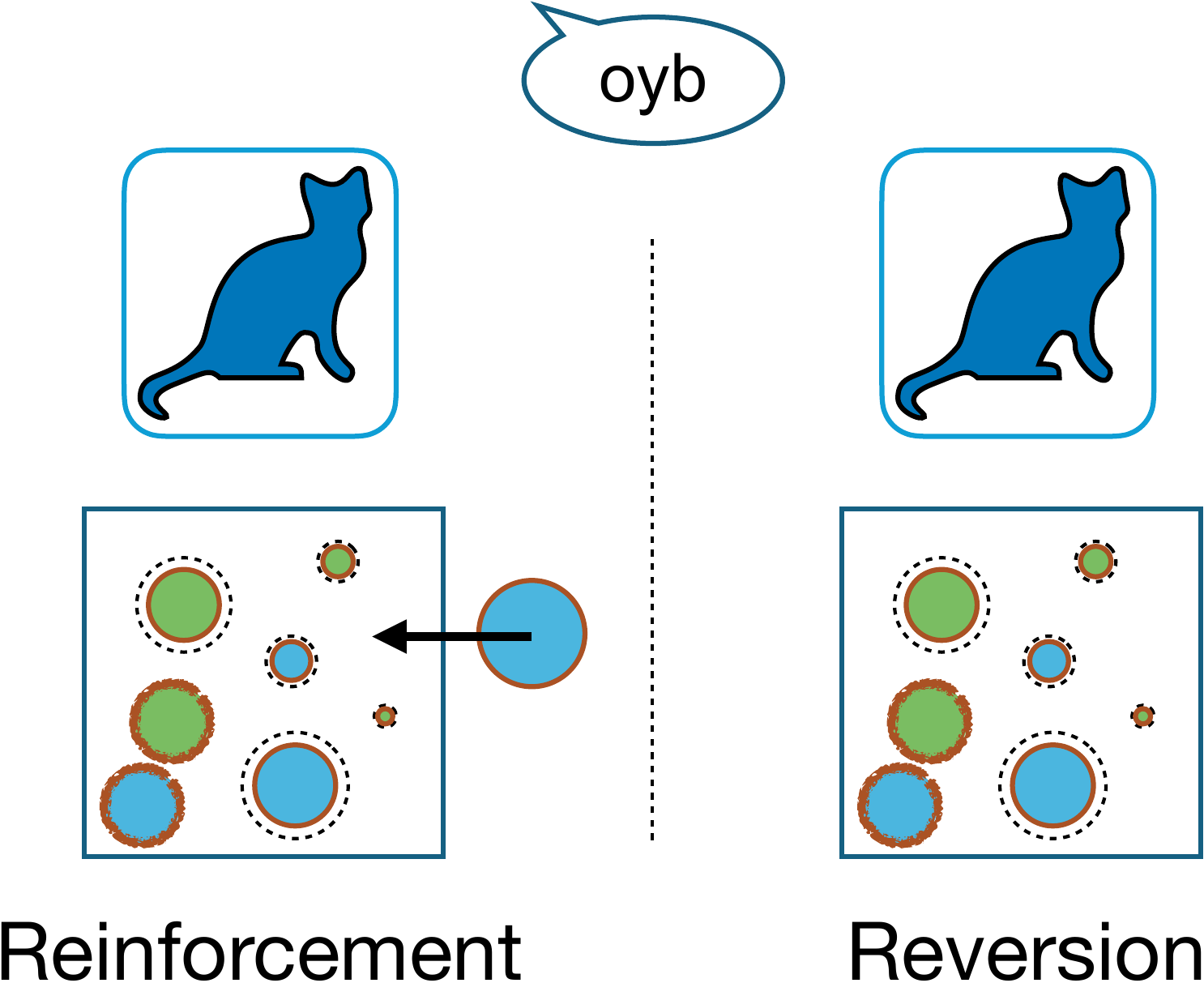}\end{center}
\caption{{\bf Retention of an interaction in memory} After the receiver has drawn an inference (here, cat) given the signal they have encountered (here, the vocalisation `oyb'), a memory of the interaction is retained. The default is to reinforce the association between the signal and interpreted meaning, operationalised by shrinking the size of memories from earlier interactions, and inserting a unit-sized memory corresponding to the signal (here, a blue circle). When feedback is available, an alternative is to decline to store the memory due to a mismatch between the interpretation and the intended topic. Previous memories shrink, yielding to a reversion to the uniform distribution (right) which occurs because prior knowledge corresponds to circles of fixed equal size (heavy outlines) in this representation.}
\label{fig:memory}
\end{figure}

To interpret the update rule (\ref{update}), it is helpful to represent it pictorially---see Fig~\ref{fig:memory}. The prior is represented by the circles with thick outlines. Since the prior represents the information available to an agent before they learn from the signallers they interact with, we assume that this cannot be forgotten, and thus these circles retain their initial size $\frac{\alpha}{S}$ at all times. The remaining circles represent memories of signals that have been interpreted with the meaning under consideration. These shrink in size each time that meaning is interpreted. In the absence of feedback, or when feedback is present and the interaction was a success, a new circle of unit size and colour corresponding to the signal that was produced is entered, thus reinforcing the association between meaning and signal. Only when we assume that communicative feedback is already present and the interaction was a failure is nothing entered into memory: this results in a reversion to the prior uniform distribution over signals.

Similar update rules have been applied in related works, albeit without the Bayesian and information-theoretic underpinnings that we have taken time to emphasise here. For example, a linear combination of new and existing behaviour, subject to linear biases, was advanced in \cite{Baxter2006} on the basis of simplicity and yields the same update when the topic was tightly constrained to a single meaning. A benefit of the current approach is that it extends to multiple arbitrarily constrained meanings with no further assumptions. Likewise, \cite{Spike2017} presents an urn model similar to that shown in Fig~\ref{fig:memory}, albeit with an ad-hoc deletion rule to model memory loss, rather than reversion to the prior. More generally, the first term in (\ref{update}) corresponds to a classic reinforcement learning update \cite{Roth1995}. In this framework, failing strategies have sometimes been suppressed by applying a negative reinforcement, i.e., inverting the sign of the first term. However, this is problematic, because it can lead to the creation of negative probabilities, which has to be prevented in some way  \cite{BerebyMeyer1998}. Here, reversion to the prior has the same effect as suppressing failing strategies, but without such problems arising. We further note that the underlying Bayesian computation is in principle highly computationally demanding, because to implement it fully would require agents to maintain probability distributions over all possible combinations of signal frequencies. This feature sometimes draws criticism of Bayesian inference as a model for human cognition \cite{Griffiths2010}. Here, we see that the algorithm for updating the posterior predictive distribution requires agents \emph{only} to keep track of association strengths between each signal and meaning and to apply the linear rule (\ref{update}).

\medskip

For reference in the following, we summarise the key parameters and variables in the model in Table~\ref{parms}.

\begin{table}[!ht]
\centering
\caption{\textbf{Parameters and variables in the model of signal emergence.}}
\begin{tabular}{|l|l|}
\hline
\multicolumn{2}{|l|}{\textbf{Parameters}}\\ \hline\hline
$N$ & Number of agents \\ \hline
$M$ & Number of meanings that can be expressed \\ \hline
$S$ & Number of signals that can be produced \\ \hline
$\rho_\ell(m)$ & Attention paid to meaning $m$ by agent $\ell$ \\ \hline
$C$ & Certainty in the attention distributions \\ \hline
$A$ & Alignment of attention between agents \\ \hline
$\lambda$ & Memory decay rate \\ \hline
$\alpha$ & Strength of the prior distribution over signals \\ \hline
\multicolumn{2}{|l|}{\textbf{Variables}}\\ \hline\hline
$\phi_\ell(s|m)$ & Frequency agent $\ell$ uses signal $s$ for meaning $m$ \\ \hline
$\psi_\ell(m|s)$ & Probability agent $\ell$ interprets signal $s$ as meaning $m$ \\ \hline
$G$ & Communicative gain of the emergent signalling system \\ \hline
\end{tabular}
\label{parms}
\end{table}

\subsection{Communicative gain}

In the initial condition, agents use every signal with equal probability for every meaning: $\phi_\ell(s|m)=\frac{1}{S}$ for all $\ell$, $s$ and $m$ because the prior distribution in the Bayesian procedure for estimating signal frequencies is invariant under exchange of signals. Such a state is non-communicative, as the signalling behaviour is identical no matter what the topic. To assess whether communication has emerged as agents repeatedly interact through the sequence of steps described above, we require a measure of communicative gain.

One possibility is the frequency with which the receiver's inference matches the signaller's topic. However, these can be the same for reasons incidental to communication. An extreme case of this is where both agents are focussed on the same meaning in every interaction (i.e., $C=A=1$). Then, the receiver will always correctly identify the topic even in the non-communicative state.

To this end, we use instead the probability $p_s$ of a \emph{blind success}, that is, whether two agents picked uniformly at random (no matter the actual social network structure) are able to communicate a topic that is also chosen uniformly (again no matter the actual distribution of attentional weights). This measure, defined as 
\begin{equation}
p_s = \frac{1}{N(N-1)} \sum_{i \ne j} \sum_{m} \frac{1}{M} \sum_s \frac{\phi_i(s|m) \phi_j(s|m)}{\sum_\mu \phi_j(s|\mu)} \;,
\end{equation}
therefore assesses the potential of the communication system to resolve ambiguity in even the most challenging situation. 

It is helpful when considering simulation results to scale this success probability such that, for a given number of meanings $M$ and signals $S$, the non-communicative state maps to $0$ whilst its maximum value of $\frac{S}{M}$ maps to $1$. We call this scaled measure the 
\emph{communicative gain} $G$:
\begin{equation}
\label{G}
G = \frac{Mp_s -1}{S-1} \;.
\end{equation}

We note that there are other possible metrics that might better capture the structure and information content of the signalling system that emerges, for example, those based on such information-theoretic measures as Kullback-Leibler divergence or mutual information. Since here we focus mostly on the simplest possible cases (such as all meanings having equal probability), communicative gain is sufficient for our needs.

\section{Results}

We now determine how and when communication systems emerge under three different assumptions as to how candidate meanings for a signal are constrained in an interaction. First, we examine the case where constraints on a signal are so tight that the topic can be identified by the receiver in every interaction. Then, we turn to the opposite extreme, where there are no such constraints and the topic can be determined only by resorting to feedback. Finally, we consider the case of greatest interest, where constraints are weak but the alignment between agents is sufficiently high that communication can nevertheless emerge in the absence of feedback.

\subsection{Tight constraints through prior expectations}

As noted above, when the certainty $C$ and alignment $A$ both equal $1$, the constraints on candidate meanings are so tight that both the signaller's and receiver's attention is always focussed on a single common meaning. This could occur for example because some unexpected and dramatic event (like a clap of thunder) prompted the interaction, or through the application of a mutual-exclusivity constraint that has ruled out all other possible meanings for the signal that has been produced \cite{Markman1988,Reisenauer2013}.

\begin{figure}[h!]
\includegraphics[width=\linewidth]{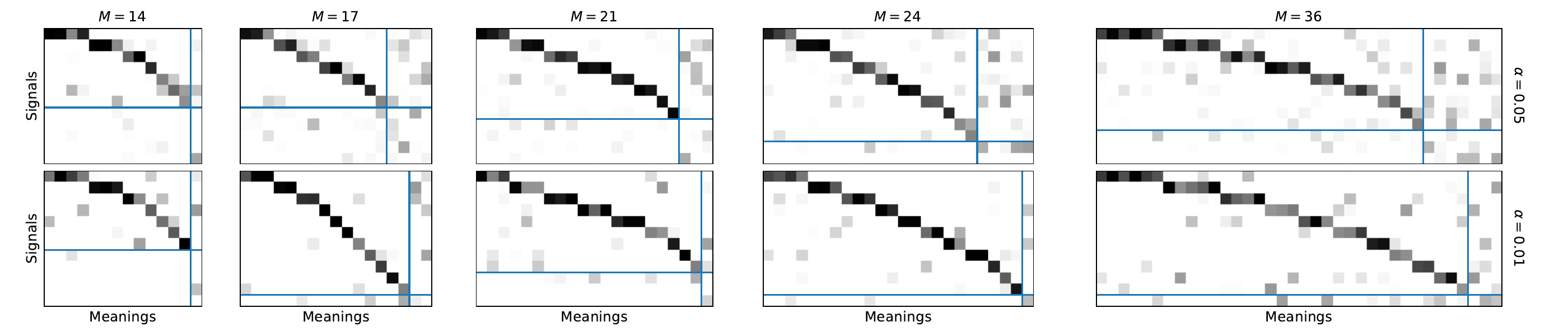}
\caption{{\bf Communication systems that emerge under tight meaning constraints.}
Patch shading indicates the frequency $\phi(s|m)$ with which each signal $s$ is used to convey meaning $m$, averaged across a society of $N=5$ agents. In all cases $\lambda=0.01$, $C=1$, $A=1$ and $S=12$. Meanings increase from $M=14$ (leftmost column) to $M=36$ (rightmost column). Prior strength $\alpha=0.05$ (upper row) or $\alpha=0.01$ (lower row). A single signal dominates for a given meaning when $\phi(s|m)>\frac{1}{2}$. Signals have been ordered by the number of meanings for which they dominate; and meanings ordered so that they are adjacent when the same signal dominates. The horizontal line indicates the boundary between those signals that dominate at least one meaning and those that do not dominate any meaning. The vertical line indicates the boundary between meanings that are dominated by a signal and those that are not. We see that more signals dominate as the prior strength is reduced.}
\label{fig:fullinfogr}
\end{figure}

Simulations of the model under these conditions tend to exhibit communication systems in which only some meanings have a dominant signal, that is one that is used with a frequency $\phi(s|m)>\frac{1}{2}$ across the society. Furthermore, there are some signals that do not dominate any meanings. Fig~\ref{fig:fullinfogr} shows how these dominant signals are distributed in single realisations of the emergence process for various combinations of the prior strength $\alpha$ and the number of meanings $M$. Such communication systems have a suboptimal communicative gain, because some signals are effectively unused. A similar result was reported by \cite{Fontanari2011} who further observed that the success rate achieved is close to that found if one assumes that each meaning is expressed by a single randomly-assigned signal.

We can explain this finding by deriving a stochastic differential equation for an agent's frequency estimate $\phi_{\ell}(s|m)$ from the update rule (\ref{update}). In the Appendix, we show that the resulting equation is
\begin{equation}
\label{migmut}
\dot\phi_{\ell}(s|m) = \frac{\lambda \rho(m)}{1+\lambda\alpha} \bigg( \left[ \phi(s|m) - \phi_{\ell}(s|m) \right] + 
\lambda\alpha \left[\frac{1}{S} - \phi_{\ell}(s|m)\right] \bigg) + \eta(t) \;,
\end{equation}
where $\eta(t)$ is a random function that models fluctuations around the average behaviour given by the first term. In this equation $\phi_\ell(s|m)$ is the frequency of a signalling strategy for agent $\ell$, whereas $\phi(s|m)$ is the corresponding frequency averaged across the society of interacting agents. This equation is valid when the set of signallers that any receiver interacts with is representative of that wider society, which seems to be true of social networks that have a sufficiently high density of long-range connections (see Appendix).

The significance of this equation is that it can be recognised as describing the dynamics of Wright's island model from population genetics \cite{Wright1931}. This is a neutral model \cite{Kimura1985}, in which the first term in square brackets describes passive migration between an island (which here corresponds to a single agent) and a mainland (here, the rest of the society). The second term describes a symmetric mutation process at a rate proportional to $\lambda\alpha$, that is, the product of the forgetting rate $\lambda$ and the prior strength $\alpha$. This mutation originates in prior expectations never being forgotten over time as learned signalling behaviour is. The stochastic term $\eta(t)$ can be identified with genetic drift.

The long-term behaviour of this population dynamics is well understood \cite{Baxter2006,Burden2018}. At high mutation rates, one typically finds coexistence between the different genotypes, which here corresponds to the non-communicative state in which all signals are used for every meaning. At low mutation rates, one signal is expected to dominate each meaning across the entire society, as a result of fluctuations arising from sampling (i.e., the cultural analogue of genetic drift). However, there are no interactions between different meanings, meaning that each signal that dominates does so independently of the others. This lack of structure, and therewith under-utilisation of the signals, is a direct consequence of the lack of ambiguity---the receiver always knows the topic ahead of signal production, so signalling cannot add any information to what is already available.

\begin{figure}[h!]
\includegraphics[width=\linewidth]{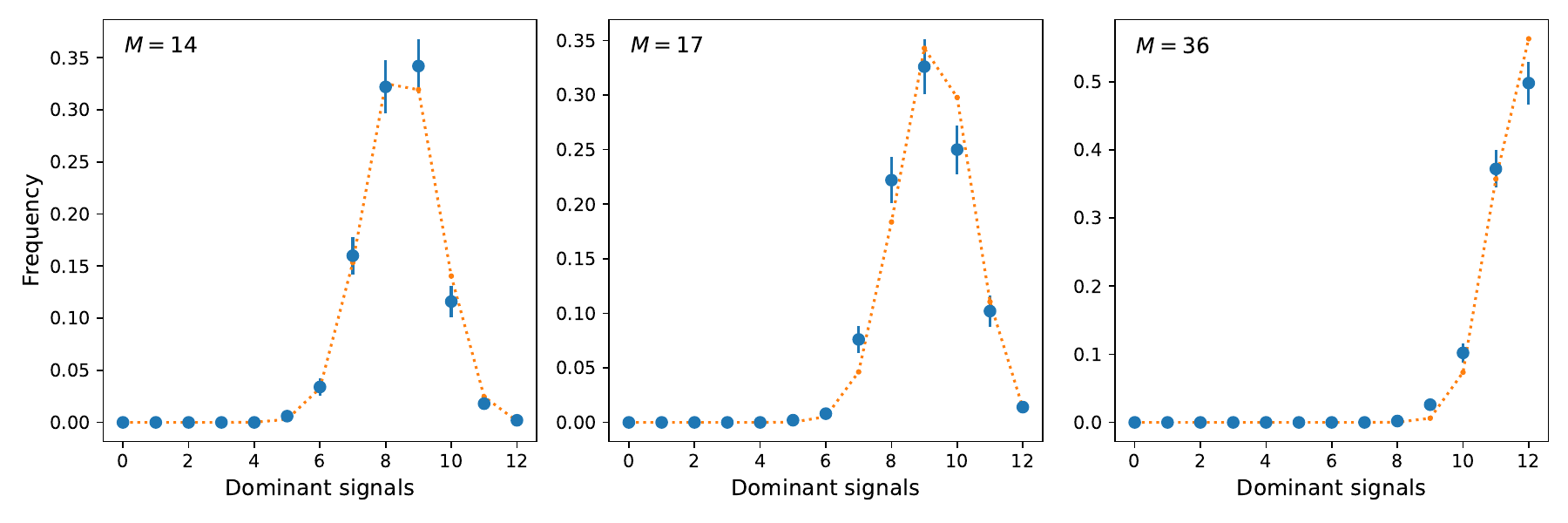}
\caption{{\bf Distribution of the number of dominant signals under tight constraints.} Empirical distribution (points) obtained from simulations with $\lambda=0.01$, $A=C=1$, $S=12$ and $\alpha=0.01$ with $M=14$ (left), $M=17$ (middle) and $M=36$ (right). The theoretical distribution is given by Eq.~(\ref{PU}), and corresponds to the case where a randomly-assigned signal dominates each meaning. We find good agreement with the theoretical prediction.}
\label{fig:unuseddist}
\end{figure}

Therefore, as $\alpha\to0$, we should expect to find the state proposed by \cite{Fontanari2011}, that is, where the probability that $D$ different signals dominate in the state that emerges is distributed as
\begin{equation}
\label{PU}
\mathbb{P}(D) = \binom{S}{D} \sum_k \binom{D}{k} (-1)^k \left(\frac{D-k}{S}\right)^M \;.
\end{equation}
We test this prediction by comparing with simulations at $\alpha=0.01$ in Fig~\ref{fig:unuseddist} for different $M$ at $S=12$. The good agreement confirms that when meanings are tightly constrained, unstructured associations between signals and meanings arise through neutral cultural evolutionary forces that are analogous to migration, mutation and genetic drift.

\subsection{Weak constraints modulated by feedback}

We now turn to the opposite extreme, where agents' attention is maximally unconstrained. This corresponds to a certainty $C=0$, implying that all agents distribute their attention across the $M$ meanings in exactly the same way in every interaction. The value of alignment parameter $A$ is irrelevant in this case. The only way for the resulting ambiguity to be resolved is by allowing communicative feedback into the model. Recall that this is implemented as shown in Fig~\ref{fig:memory}, where reinforcement occurs only after successful interactions, and reversion to the initial uniform distribution over signals occurs on failure. Simulations show that a communicative state that utilises all signals is typically arrived at, as long as the product of the forgetting rate $\lambda$ and prior strength $\alpha$ is sufficiently small.

We can develop a quantitative understanding of the communicative gain that is achieved when this occurs by deriving the stochastic differential equation for an agent's frequency estimate $\phi_{\ell}(s|m)$. As previously, the procedure is described in the Appendix. This time, we arrive at a replicator-mutator type equation \cite{Nowak2006}
\begin{equation}
\label{repmut}
\dot\phi_{\ell}(s|m) = \frac{\lambda \rho(m)}{\beta_{\ell}(m)+\lambda\alpha} \bigg( \phi_{\ell}(s|m) \left[ f_{\ell}(s|m) - f_{\ell}(m) \right] + \lambda\alpha \left[\frac{1}{S} - \phi_{\ell}(s|m)\right] \bigg) + \eta(t) \;.
\end{equation}
Here, $f_{\ell}(s|m)$ is the local \emph{fitness} of signalling strategy $s$ when $m$ is the receiver's interpretation, $f_\ell(m)$ is the mean fitness over all signals competing to express meaning $m$ and $\beta_{\ell}(m)=f_\ell(m)$. As above, the second term in square brackets corresponds to symmetric mutation, and $\eta(t)$ to genetic drift. This is not a neutral model, but one in which signalling frequencies are driven by selection. Broadly speaking, the signalling strategy with the highest fitness will proliferate, until balanced by the mutation that derives from forgetting. 

In the current scenario, we find in the Appendix that the fitness is $f_\ell(s|m) \approx \psi(m|s)$, the probability that a randomly-chosen receiver interprets signal $s$ as meaning $m$, as defined by Eq.~(\ref{psi}). That is, competition within each local population favours the signal which is most likely to be correctly interpreted. This can be viewed as a form of higher-order reasoning, implemented as an obverter strategy \cite{Oliphant1997,Spike2017} or recursively within the rational speech act framework \cite{Goodman2016}. Here, this is not built into the model, but is an emergent product of an interaction between agents behaving predictably and communicative feedback.

\begin{figure}[h!]
\begin{center}\includegraphics[width=0.5\linewidth]{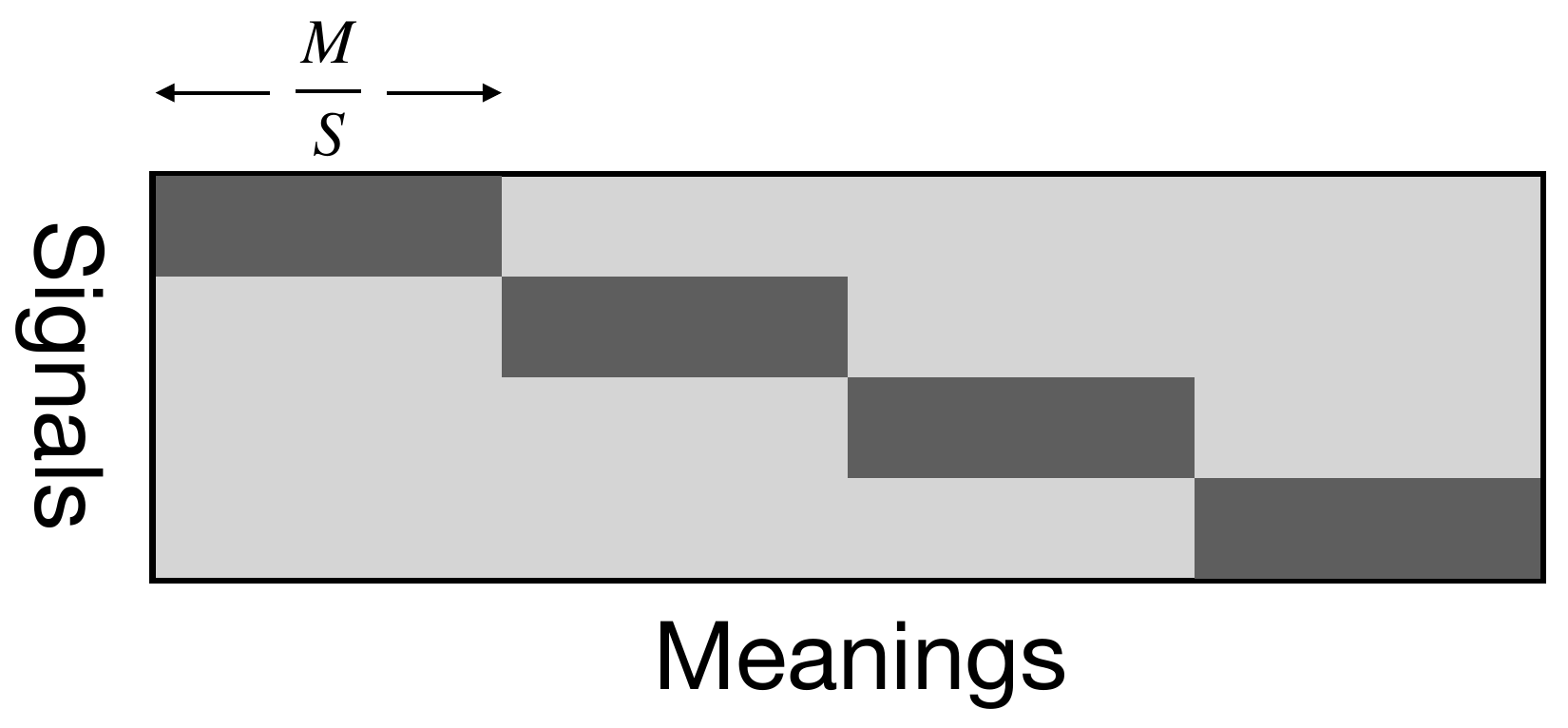}\end{center}
\caption{{\bf Symmetrically-structured communication system.} In a symmetrically-structured communication system, each signal is used with frequency $x$ for exactly $\frac{M}{S}$ meanings that are not shared with any other signals (shown dark shaded). The remaining signals (light shaded) are used with equal frequency $\frac{1-x}{S-1}$.}
\label{fig:sym}
\end{figure}

The full high-dimensional system of equations (\ref{repmut}) does not facilitate a straightforward analysis. However, to understand the emergence of communication it is sufficient to consider the special case where the attention distribution is uniform, $\rho(m) = \frac{1}{M}$, and the single-coordinate reduction that is obtained if we restrict the dynamics to a space of symmetrically-structured communication systems, illustrated in Fig~\ref{fig:sym}. These communication systems partition the meanings into $S$ disjoint subsets of equal size $\frac{M}{S}$, each conveyed by a common signal with the same frequency $x$. The other signals are used for each of those meanings with frequency $\frac{1-x}{S-1}$. Such systems are optimal in the sense that they distinguish between different meanings to the maximum extent possible (given the number meanings and signals and the effect of forgetting).

In this symmetric case, we find in the Appendix that the deterministic part of Eq.~(\ref{repmut}) simplifies to
\begin{equation}
\label{xdotf}
\dot{x} = \frac{\lambda}{\beta(x) + \lambda\alpha} \frac{1}{M} \left[ \frac{\Gamma S^2}{S-1}  x(1-x) - \lambda\alpha \right] \left[x-\frac{1}{S}\right] \;,
\end{equation}
where the present case of full ambiguity and feedback-driven learning, the parameter $\Gamma=\frac{1}{M}$. (We will obtain the same equation, but with a different value of $\Gamma$ when we examine the case without feedback below.)

\begin{figure}[h!]
\begin{center}\includegraphics[width=\linewidth]{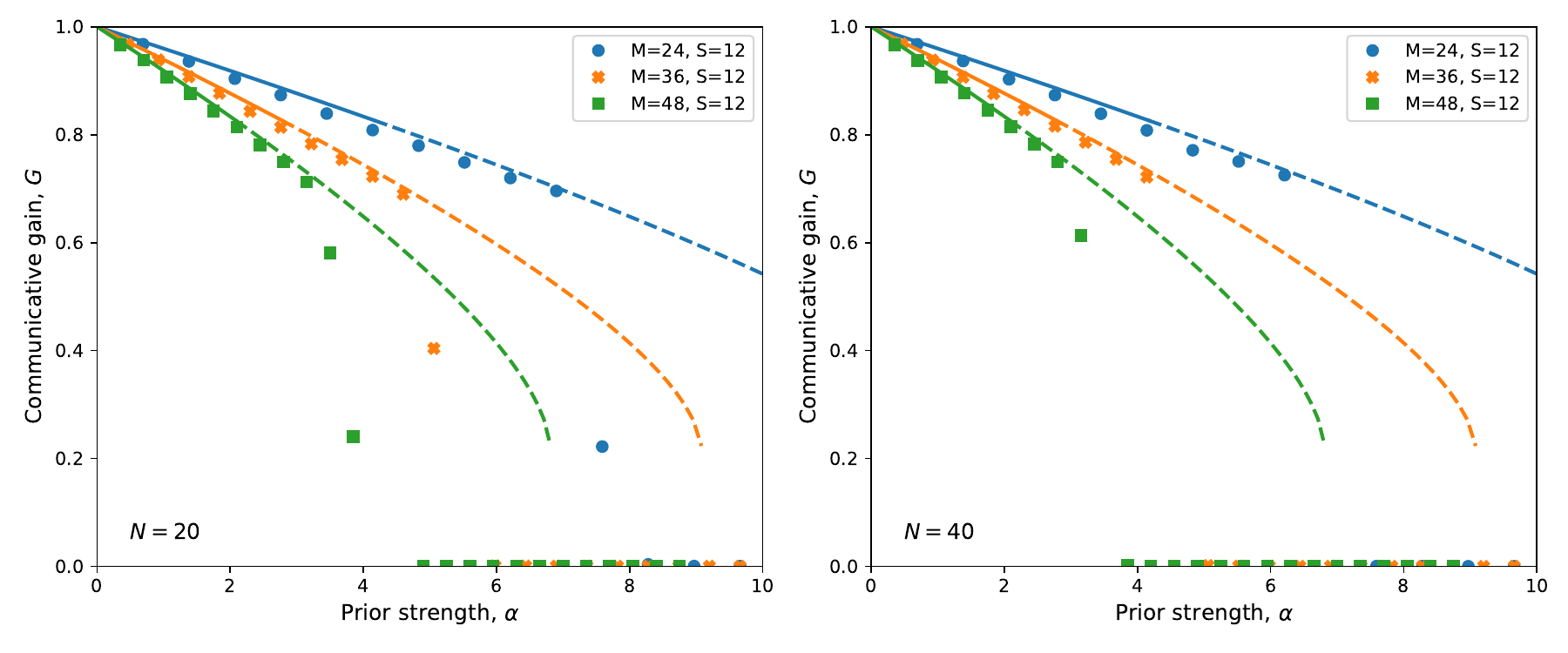}\end{center}
\caption{{\bf Communicative gain with feedback.} Lines show the communicative gain $G$ expected for different prior strengths $\alpha$ at the communicative fixed point defined by (\ref{xc}) when selection is driven by feedback and interactions are maximally ambiguous. The solid lines correspond to where the non-communicative initial condition is unstable, and the dashed lines where it is stable. Points are from simulations with $C=0$, $A=1$, $\lambda=0.01$, $S=12$ signals, $M=24,36$ or $48$ meanings in a society with $N=20$ agents (left) and $N=40$ agents (right).}
\label{fig:fb}
\end{figure}

We see that the non-communicative state where $x=\frac{1}{S}$ is a fixed point of this equation. It is unstable when $\lambda\alpha<\Gamma$, which means any fluctuation away from the initial condition will grow, and the system will evolve deterministically towards a different fixed point. The fixed point of interest is the communicative state where
\begin{equation}
\label{xc}
x = \frac{1}{2}\left( 1 + \sqrt{1-\frac{4\lambda\alpha (S-1)}{\Gamma S^2}} \right) \;.
\end{equation}
In Fig~\ref{fig:fb}, we plot the communicative gain (\ref{G}) that is expected at this fixed point as a function of the prior strength $\alpha$ at fixed $\lambda=0.01$, and find that when $\lambda\alpha < \frac{1}{M}$, the system does indeed evolve into this communicative state.

The situation above this threshold is more subtle. The communicative state exists up to $\lambda\alpha < \frac{\Gamma S^2}{4(S-1)}$, and is stable whenever it exists. Thus, there is a region where both the non-communicative and communicative states are stable, shown with the dashed lines in Fig~\ref{fig:fb}. Here it may be possible to reach one state from the other through a sufficiently large fluctuation, and we find a transition to communication occurs for lower values of $\lambda\alpha$ in this intermediate range. The left and right panels of Fig~\ref{fig:fb} shows the state that is reached at two different society sizes ($N=20$ and $N=40$ agents, respectively). We see that the region in which a fluctuation allows the non-communicative state to be escaped is slightly smaller in the larger society. We therefore expect that the probability of a fluctuation large enough to reach the communicative state will vanish as the society size increases, and that in the limit of an infinite society, communication would emerge only when the non-communicative state is unstable, i.e., when $\lambda\alpha<\frac{1}{M}$.

An intriguing consequence of the way that feedback is implemented in this model is that for fixed cognitive parameters (i.e., forgetting rate $\lambda$ and prior strength $\alpha$), there is an upper limit on the number of meanings that agents can entertain if communication is to emerge. The origin of this limit is the reversion to uniform distribution over signals when communicative failure occurs. Since the probability of a failure is high in the non-communicative state when the number of meanings $M$ is large, any fluctuation towards systematic use of one signal over the others for any given meaning is quickly suppressed. This is the case even with just two signals, that is, if agents are faced with the task of dividing a large meaning space into two categories.

\subsection{Weak constraints modulated by shared intentionality}

We now tackle the case where candidate meanings are weakly constrained (i.e., $C$ close to zero) and we do not allow communicative feedback. Now, the only way that communication could emerge is if receivers are able to exploit shared intentionality, i.e., correlations between a signaller's behaviour and how their own attention is distributed. In the Appendix, we show that the stochastic differential equation that governs the evolution also takes the form of the replicator-mutator equation (\ref{repmut}), albeit with $\beta_{\ell}(m)=1$ and the very different local fitness
\begin{equation}
\label{xslf}
f_{\ell}(s|m) = 1 + \frac{C [ A \phi(s|m) - \phi_{\ell}(s|m) ]}{\phi(s)} + \frac{C(1-A) \sum_\mu \rho(\mu) \phi(s|\mu)^2}{\phi(s)^2} \;.
\end{equation}
Here $\phi(s) = \sum_m \phi(s|m) \rho(m)$ is the frequency with which signal $s$ is produced over all interactions across the society.

From this expression we first observe that if $C=0$, all signalling strategies have the same fitness. Thus, in this case, symmetric mutation and genetic drift operate independently for each agent, meaning that no shared communicative behaviour can emerge. This is the expected result, as there is no variation in attention that can be exploited here. 

To understand what happens for small nonzero $C$, we first assume that individual differences between agents can be neglected: that is, one can take $\phi_{\ell}(s|m)=\phi(s|m)$. When, as previously, we restrict to the space of symmetrically-structured communication systems, an equation of the form (\ref{xdotf}) results with $\beta(x)=1$ and the parameter $\Gamma = -C(1-A)$ which cannot be positive since $C>0$ and $A\le1$. Recalling that the non-communicative fixed point is unstable if $\lambda\alpha < \Gamma$, we conclude that there is no possibility of communication emerging spontaneously within this analysis.

\begin{figure}[h!]
\begin{center}\includegraphics[width=\linewidth]{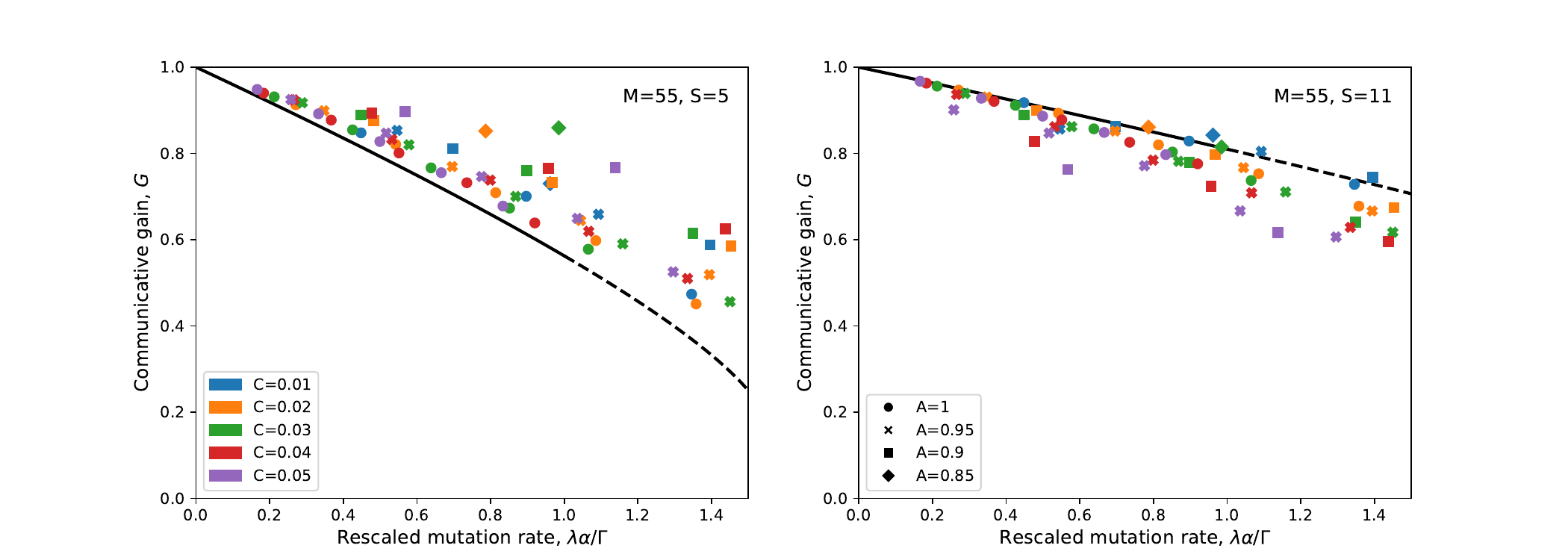}\end{center}
\caption{{\bf Communicative gain in the absence of feedback.}  The location of the communicative fixed point (\ref{xc}), and hence the gain in the communicative state, is predicted to depend on the ratio $\lambda\alpha/\Gamma$ where $\Gamma$ is the threshold value (\ref{thres}). Lines show this predicted gain for $S=5$ (left) and $S=11$ (right), with the solid part of the line indicating where the non-communicative fixed point is unstable. We find that simulation data (points) for various combinations of $C$ (denoted by colours) and $A$ (by marker shape) are in reasonable agreement with these predictions. In all simulations $M=55$ and $\lambda=0.01$.}
\label{fig:ob}
\end{figure}

This, however, is not what happens in simulations. For example, Fig~\ref{fig:ob} clearly shows that a nonzero communicative gain is possible for a wide range of parameter combinations. We now show that communication is selected for deterministically in an arbitrarily large society, and therefore not the result of the type of fluctuation that only occurs in small systems. Rather, it is the consequence of individual differences between agents that are always present. To this end, it is helpful to appeal first to the simplest possible case, where the system comprises two agents whose attention is always fully aligned ($A=1$) and there are just two signals available. Further assuming, as in the symmetrically-structured states, that the overall frequency of each signal $\phi(s)$, averaged across all meanings, remains at $\frac{1}{2}$ throughout the evolution, the local fitness (\ref{xslf}) simplifies to
\begin{equation}
f_{\ell}(s|m) = 1 + 2C [\phi(s|m) - \phi_{\ell}(s|m)] \;.
\end{equation}
Since the frequencies are normalised, $\sum_s \phi_{\ell}(s|m)=1$, the mean local fitness is
\begin{equation}
f_{\ell}(m) = 1 + 2C [2\phi_{\ell}(s|m) - 1] [\phi(s|m) - \phi_{\ell}(s|m)]  
\end{equation}
and the fitness difference that appears in the replicator-mutator equation is
\begin{equation}
f_{\ell}(s|m) - f_{\ell}(m) = 4C [1 - \phi_{\ell}(s|m)] [\phi(s|m) - \phi_{\ell}(s|m)] \;.
\end{equation}

We now focus on the signal $s^*$ that is used most frequently for the meaning $m^*$ under consideration at the society level, that is, with a frequency $\phi \equiv \phi(s^*|m^*) \ge\frac{1}{2}$. We further identify the agent who is using that signal with a frequency that is at least as high as this average, and denote them with a $+$ sign. This frequency we write as $\phi_+ \equiv \phi_{+}(s^*|m^*) = \phi + \epsilon$ where $\epsilon\ge 0$. The deterministic part of the replicator-mutation equation (\ref{repmut}) for this agent in this special case is
\begin{equation}
\label{dot+}
\dot{\phi}_+ = \frac{\lambda }{1+\lambda\alpha} \frac{1}{M} \left[ - 4C\epsilon\phi_{+}(1-\phi_{+}) + \lambda\alpha\left( \frac{1}{2} - \phi_+ \right) \right]
\end{equation}
Similarly, for the other agent, who uses signal $s^*$ for meaning $m^*$ less often, it is
\begin{equation}
\label{dot-}
\dot{\phi}_- = \frac{\lambda }{1+\lambda\alpha} \frac{1}{M} \left[ 4C\epsilon\phi_{-}(1-\phi_{-}) + \lambda\alpha\left( \frac{1}{2} - \phi_- \right) \right] \;.
\end{equation}

\begin{figure}[h!]
\begin{center}\includegraphics[width=0.5\linewidth]{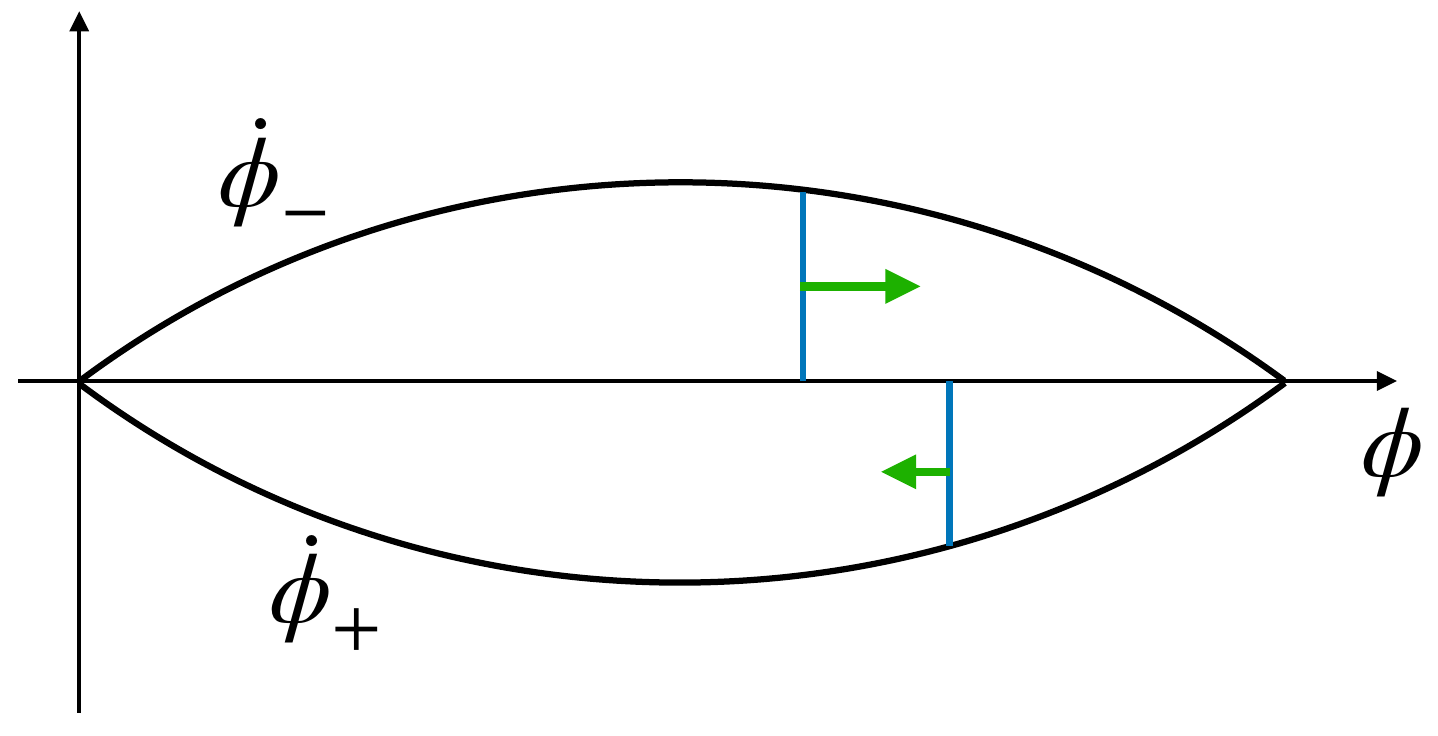}\end{center}
\caption{{\bf Interaction between competition and individual variability.}  The rate at which the lower signal frequency among the two members of the society grows is given by the upper curve, while that for the higher signal is given by the lower curve. When the mean of these two frequencies exceeds $\frac{1}{2}$, the lower signal frequency increases faster than the higher frequency decreases (shown by the arrows), yielding a net increase for this majority signal.}
\label{fig:comp}
\end{figure}

When the mutation rate $\lambda\alpha$ is very small, the dynamics are dominated by the selection term (i.e., the one proportional to $C$. Since the signal $s^*$ is by definition in the majority, $\phi\ge\frac{1}{2}$, then for small deviations $\epsilon$ from this average we have that the rate at which $\phi_{+}$ shrinks through negative selection is smaller than the rate at which $\phi_{-}$ grows through positive selection---see Fig~\ref{fig:comp}. Thus, the combined effect of this competition is for the mean frequency $\phi=\frac{\phi_++\phi_-}{2}$ to grow overall, i.e., a net positive selection.  This selective effect, however, is countered by mutation. When one accounts for both evolutionary forces, one finds that the mean frequency, obtained by summing (\ref{dot+}) and (\ref{dot-}) and dividing by two, obeys the equation
\begin{equation}
\dot{\phi} = \frac{\lambda }{1+\lambda\alpha} \frac{1}{M} \left[ 8C\epsilon^2- \lambda\alpha \right] \left[  \phi - \frac{1}{2} \right] \;.
\end{equation}
In words, this equation states that if the variance in signalling frequencies over different agents in the society, $\epsilon^2$, exceeds $\frac{\lambda\alpha}{8C}$, the signal that is in the majority for the meaning under consideration will tend to grow in frequency. Importantly, the net positive selection is driven by individual differences that arise spontaneously as a result of the randomness inherent in the interactions between agents, and whose magnitude does not depend on the size of the overall society. Therefore, although this deterministic instability is created by a fluctuation effect, it is not the rare type of fluctuation that allows stable fixed points to be escaped in small societies, and is suppressed in large ones.

In the Appendix we extend the analysis to the general case of more than two agents and signals, obtaining an expression for the fitness that applies at the society level, rather than the individual level. Using this expression in (\ref{repmut}), and restricting again to symmetric communication systems, we obtain Eq.~(\ref{xdotf}) with the threshold $\Gamma$ now given by
\begin{equation}
\label{thres}
\Gamma = C(1-V)\left(A - \frac{1-V}{1+V}\right) \;.
\end{equation}
Here, $V$ quantifies variability in signal frequencies for a fixed meaning across the society in the same way that the certainty $C$ quantifies variation in attention weights over interactions, that is, analogously to (\ref{C}),
\begin{equation}
\label{V}
V = \frac{\sum_s {\rm Var}[\phi_\ell(s|m)]}{1 - \sum_s \mathbb{E}[\phi_{\ell}(s|m)]^2}
\end{equation}
where both the expectation value and variance are over agents $\ell$. In common with $C$, this parameter lies in the range $0\le V \le 1$, and is zero if all agents are identical in their signalling of meaning $m$ and one if they are maximally different, given the society averages. In the Appendix we further provide an estimate of the value of $V$ that applies when we take into account the amplitude of the drift term in the replicator-mutator equation (\ref{repmut}).

The key point is that if both the alignment $A$ and variability $V$ are sufficiently large, the requirement that $\lambda\alpha < \Gamma$ for communication to emerge spontaneously can be satisfied. The threshold (\ref{thres}) does not depend explicitly on the number of agents $N$, meanings $M$ or signals $S$, and the implicit dependence on $M$ through $V$ can be neglected when $M$ is large (see Appendix). Thus, there is a deterministic instability to bootstrapping communication in populations of arbitrary size, no matter how many meanings or signals are available, as long as interacting agents' attention covaries sufficiently strongly from one interaction to the next, and differences between individuals are able to grow.

To test these predictions, we note from (\ref{xc}) that the location of the communicative fixed point depends on the number of signals $S$ and the scaled mutation rate $\frac{\lambda\alpha}{\Gamma}$. Thus, we should find for different combinations of $M$, $A$ and $C$ (and hence $V$, which depends on all of these quantities), that the communicative gain falls along a single curve when plotted as a function of $\frac{\lambda\alpha}{\Gamma}$ for fixed $S$. Simulation data in Fig~\ref{fig:ob} provide reasonable support for this prediction, with the deviations most likely arising from imprecision in the estimate of $V$ presented in the Appendix.

\begin{figure}[h!]
\begin{center}\includegraphics[width=0.75\linewidth]{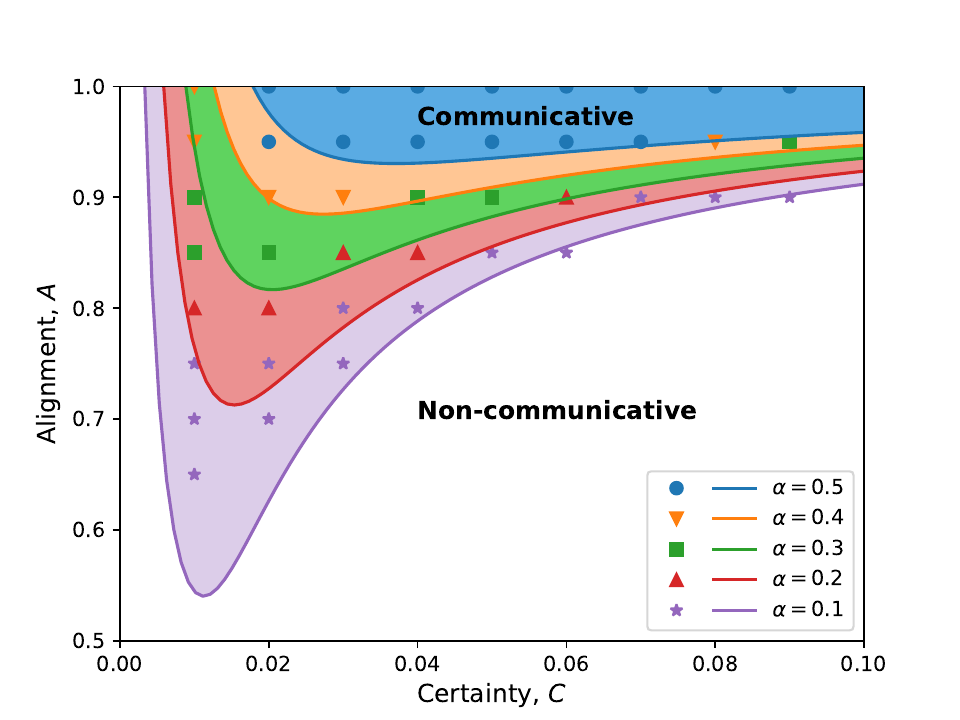}\end{center}
\caption{{\bf Phase diagram for the emergence of communication.} Shaded regions show combinations of $C$, $A$ and $\alpha$ that facilitate communication without feedback in the limit of an infinite number of meanings $M$. Simulations are for the case $\lambda=0.01$, $S=11$, $M=55$ and points are plotted when the gain $G>0.5$.}
\label{fig:pd}
\end{figure}

Moreover, we can identify combinations of $A$ and $C$ where communication is expected to arise when other parameters are held fixed. These are compared in Fig~\ref{fig:pd} with regions where a communicative gain exceeding one half is found in simulation. Strictly speaking, the region determined analytically should coincide with that where any positive gain is achieved. However, that region is found to be larger than that predicted (although has the same qualitative shape). We suspect that this is a consequence of finite-size fluctuations that allow the communicative state to be reached even when the non-communicative state is stable also being present. The existence of these is suggested by Fig~\ref{fig:ob} where we find points close to the dashed part of the line. To confirm this would require running the simulations in much larger societies: however, this is computationally challenging as the simulation run time required grows linearly in the size of the society $N$, and we would probably need to simulate societies at least two or three orders of magnitude larger to gain such confirmation.

\section{Discussion}

In this work, we constructed a model within which agents can learn signalling behaviour from other members of their society, correlating this with how their attention is distributed over possible meanings, and optionally utilising communicative feedback. This model was grounded in principles of Bayesian inference \cite{Griffiths2010,Gill2015} and information theory \cite{Cover2006}. An important feature of the model was the ability to control constraints on candidate meanings through the extent to which attentional weights varied between interactions and between interacting agents.

Our main finding is that effective communication can emerge spontaneously in large well-connected societies without any pre-existing ability to communicate success or failure. The necessary ingredients are: (i) that agents learn to predict which signal is used by others to express each given meaning \cite{Bar2011,Pickering2013}; (ii) act cooperatively by conforming to these predictions \cite{Tomasello2005,Chater2018,ContrerasKallens2024}; and (iii) possess sufficient shared intentionality \cite{Tomasello2007} that mental representations about likely topics are well aligned before a signal is produced \cite{Stolk2016,Schilbach2025}. Once individual differences have opened up between agents, communication, and by extension language, can be bootstrapped from social cognitive capabilities that are not specific to communication or language \cite{Tomasello2005,Tomasello2008}.

There is no limit on the number of signals, meanings or agents for bootstrapping to be possible. Agents also do not need to be equipped with strong constraints on the structure of the communication system. Instead, it is sufficient for the probability that signaller and receiver agree on the topic in the absence of a convention to be slightly above chance. Most importantly, a threshold level of alignment must be exceeded, suggesting that species with limited shared intentionality would not be able to bootstrap a large socially-learned communication system. A curious finding is that feedback-driven learning is effective for communication only when the number of meanings is sufficiently small. We do not intend to suggest that feedback plays no role in everyday conversation or language acquisition, just that it might be counterproductive while establishing a communication system \textit{de novo}.

The inevitability of effective communication under the above conditions was missed in earlier studies for various reasons. First, if topics are highly constrained (large $C$ in the model), signals have little work to do and less effective systems that fail to utilise all signalling behaviours emerge \cite{Fontanari2011}. Second, fluctuations in small societies can be large enough that a communicative state is found even though at the deterministic level, non-communication is stable. This can be seen in Fig~\ref{fig:fb} and Fig~\ref{fig:ob}, where communicative gain was achieved in simulations where both communication and non-communication are stable (dashed lines). Such fluctuations are expected to be suppressed in large societies, consistent with earlier simulations \cite{Vogt2003}.

Although minimal by design, our model lends itself to a number of potential applications and generalisations. First,  we assumed that our agents initially had no prior exposure to linguistic behaviour, appropriate to an emergence scenario. An alternative would be to bring agents together who have created different signalling systems, thereby modelling a language contact scenario. It would be interesting to see if the model replicates features observed for example in the emergence of new sign languages \cite{Senghas2001}. It would also be interesting to understand how communication systems emerge and change when the interaction network is not fixed, but is affected by the signals adopted by individual agents. This might, for example, allow for a self-contained model of sociolinguistic effects, consistent with the view of language as a complex adaptive system \cite{Beckner2009}. Second, for reasons of mathematical tractability, we restricted to the case where all meanings are equally likely to be attended to. It would be worthwhile to generalise to non-uniform meaning distributions, including those where, for example, attending to parts of an object implies also attending to the whole object. One would then be able to determine whether the components we have included in our model are sufficient to reproduce the colexification properties that have been established for natural languages \cite{Futrell2024}.

Finally, we note that a major limitation in the present study of emergence is that the set of signals and meanings have been assumed fixed and available to agents.  In particular, this implies that all agents use all signals interchangeably in the initial condition, whereas a more natural scenario would be one in which meanings and signals are created by agents as they experience the world around them.  One way to address this shortcoming might be to exploit aspects that our model has in common with transformer models \cite{Ashish2017} in which meanings are represented as vectors and attention is distributed over meanings as the model attempts to predict the next signal in a sequence. Our results suggest that populations of transformers that are exposed to correlated views of a complex meaning space and seek to predict each others' output could bootstrap a common classification of that meaning space without relying on a pre-existing set of word embeddings or the assumption of an explicit reward system \cite{Sukhbaatar2016}. As well as providing a means to test the robustness of the mechanism for bootstrapping communication that we have identified here in less idealised scenarios, such investigations might also suggest novel self-supervised multi-agent machine learning algorithms.

\section*{Acknowledgments}
We thank Tim Rogers for comments on the manuscript. For the purpose of open access, the author has applied a Creative Commons Attribution (CC BY) licence to any Author Accepted Manuscript version arising from this submission.

\section*{Data availability}

Code is available at \cite{code}. Simulation data is available at \cite{data}.

\appendix

\section{Averaging the update rule}

The deterministic term in (\ref{migmut}) and (\ref{repmut}) is found by taking the average of the update rule (\ref{update}) over all interactions where agent $\ell$ is the receiver. Assuming each agent acts as a receiver on average once per unit time, this corresponds to the differential equation
\begin{equation}
\label{nasty}
\dot{\phi}_{\ell}(s|m) = \mathbb{E}_{k,\rho_k,\rho_\ell} \left[ \sum_{\mu,\sigma} \rho_k(\mu) \phi_k(\sigma|\mu) \psi_{\ell}(m|\sigma)  \delta\phi_{\ell}(s|m) \right] + \eta(t)
\end{equation}
Here, $\mathbb{E}_{k,\rho_k,\rho_\ell}$ denotes an expectation over the distribution of signallers $k$ and the perspectives $\rho_k$ and $\rho_\ell$ that the signaller and receiver hold in an interaction. Fluctuations around the expectation value given by the first term are represented by the function $\eta(t)$, the details of which are required only to estimate the variability parameter $V$ that is relevant when communication emerges in the absence of feedback. We further recall that $\psi_{\ell}(m|\sigma)$ is defined by the Bayesian inference procedure, Eq.~(\ref{psi}) in the main text.

Calculating the expectation value in (\ref{nasty}) is in general not straightforward, and certain assumptions are necessary to make analytical progress. These are: (i) that the number of meanings is sufficiently large that the sums over meanings in (\ref{nasty}) can be replaced by society averages; (ii) that certainty is low, $0\le C\ll 1$, implying limited variation in meaning distributions between interactions, or equivalently, weak constraints on statistical learning; and (iii) that each agent interacts with a sample of signallers that is representative of the wider society.

\section{Tight constraints}

The simplest case is where constraints are maximally tight ($C=1$). The Bayesian inference process is then deterministic, meaning that $\psi_{\ell}(m|\sigma) = \delta_{m\mu}$: that is, the receiver's interpretation $m$ always coincides with the speaker's topic $\mu$. Then, (\ref{nasty}) simplifies to
\begin{equation}
\dot\phi_{\ell}(s|m) = \mathbb{E}_{k,\rho_k} \left[ \sum_{\sigma} \rho_k(m) \phi_k(\sigma|m)  \delta\phi_{\ell}(s|m) \right] + \eta(t) \;.
\end{equation}

To obtain the migration-mutation equation (\ref{migmut}), one first averages over all topic distributions $\rho_k$, which replaces $\rho_k(m)$ with the society average $\rho(m)$. We now invoke the first of our main assumptions, which is that each receiver interacts with a representative sample of signallers. Put another way, we assume that behaviour spreads rapidly across a social network, which tends to occur when the maximum distance between any two agents across the interaction network is small \cite{Baxter2008}. This is likely appropriate for real social networks. Then, the average  $\mathbb{E}_k [\phi_k(\sigma|m)]$ over that set of signallers is exactly the society average  $\phi(\sigma|m)$. Now all that is required is to substitute in the update rule (\ref{update}) from the main text, noting that $\omega_{m\mu}=1$ always in this condition, and we arrive at Eq.~(\ref{migmut}) in the main text.

\section{Weak constraints}

We now turn to the derivation of the stochastic replicator-mutation equation (\ref{repmut}) that applies when meanings are weakly constrained ($0\le C < 1$). The mutation and drift terms will be the same as were already obtained above, so we need only focus on the first term in the update rule (\ref{update}) that corresponds to retaining memory of an interaction. Substituting this into (\ref{nasty}), we find that it contributes
\begin{equation}
\label{inter}
\frac{\lambda \rho(m)}{\beta_{\ell}(m) + \lambda\alpha} \phi_{\ell}(s|m) \left[ f(s|m) - \sum_\sigma \phi_{\ell}(\sigma|m) f(\sigma|m) \right]
\end{equation}
to the rate of change of the local frequency $\phi_{\ell}(s|m)$, where the fitness is given by
\begin{equation}
\label{finit}
f_{\ell}(s|m) = \frac{1}{\rho(m)} \mathbb{E}_{k,\rho_k,\rho_\ell} \left[ \frac{ \sum_\mu \rho_k(\mu) \phi_k(s|\mu)  }{ \sum_\nu \rho_{\ell}(\nu) \phi_{\ell}(s|\nu)  } \rho_{\ell}(m) \omega_{m\mu} \right] \;.
\end{equation}
The sum over $\sigma$ in (\ref{inter}) is the mean fitness of all signalling strategies employed by agent $\ell$, denoted $f_{\ell}(m)$ in the main test.

Recall that in the update rule the function $\omega_{m\mu} = \delta_{m\mu}$ when feedback is present, while $\omega_{m\mu} = 1$ when it is not. The procedure for evaluating the fitness is different in these two cases, and we will therefore consider them separately. We are most interested in cases where the contexts of use are highly ambiguous, that is, when the certainty $C$ is small. This means that to evaluate the fitness $f_{\ell}(s|m)$ we identify the lowest order in $C$ at which we obtain an expression for the fitness that depends on the signal $s$, and therefore contributes to the dynamics.

\subsection{With feedback}

Here, we obtain signal-dependent fitness functions at zeroth order in $C$, since feedback allows meanings to be distinguished even when every interaction is identical. Since differences between $\rho_\ell(m)$ and its mean over interactions $\mathbb{E}[\rho_{\ell}(m)] = \rho(m)$ are of order $C$, we can replace $\rho_\ell(m)$ with $\rho(m)$ to obtain the desired leading-order expression. The same goes for $\rho_k(m)$. Since the resulting expressions then no longer depend on $\rho_k(m)$ or $\rho_\ell(m)$, averaging over them is trivial, and we find
\begin{equation}
f_{\ell}(s|m) = \mathbb{E}_{k} \left[ \frac{  \rho(m) \phi_k(s|m)  }{ \sum_\nu \rho(\nu) \phi_{\ell}(s|\nu)  }  \right] \;.
\end{equation}
The simplification in the numerator arises due to the fact that $\omega_{m\mu}= \delta_{m\mu}$ under feedback-driven learning. 

To simplify the denominator, we now make use of the assumption that the total number of available meanings is large. Then,
\begin{equation}
\label{bigMapx}
\sum_\nu \rho(\nu) \phi_{\ell} (s|\nu) \approx \sum_\nu \rho(\nu) \phi(s|\nu) = \phi(s) \;,
\end{equation}
where the approximate equality holds as a consequence of the central limit theorem, as long as any correlations between fluctuations in $\phi_{\ell}(s|\nu)$ for different meanings are weak. We note that $\phi(s)$ is the frequency that signal $s$ is used at the society level. If we now further invoke the assumption that the set of signallers that receiver $\ell$ interacts with are representative of the society at large, then the average over $k$ above converts $\phi_k(s|m)$ to the society average $\phi(s|m)$. We thus then find the fitness to be
\begin{equation}
f_{\ell}(s|m) \approx \frac{\rho(m) \phi(s|m)}{\phi(s)} = \psi(m|s) \;,
\end{equation}
that is, the probability that $m$ is interpreted whenever $s$ is produced in an interaction through the Bayesian inference procedure (\ref{psi}).  This is the result quoted in the main text.

\subsection{Without feedback}

When no feedback is available, the information provided at each interaction is always recorded, and $\omega_{m\mu}=1$. If we do not account for fluctuations in the signaller's and receiver's meaning distributions, then we find that all signalling strategies have the same fitness. Therefore, we take $\rho_k(\mu) = \rho(\mu) + \delta \rho_k(\mu)$ for the signaller, and $\rho_\ell(\mu) = \rho(\mu) + \delta \rho_\ell(\mu)$ for the receiver, noting that the differences from the metapopulation values $\rho(\mu)$ may be correlated due to signaller and receiver having some degree of alignment between their perspectives. 

A generic model of small fluctuations can be constructed by drawing the meaning frequencies from a Dirichlet distribution such that $\mathbb{E}_{\rho_k} [ \delta \rho_k(\mu) ] = \mathbb{E}_{\rho_\ell} [ \delta \rho_\ell(\mu) ] = 0$ and
\begin{equation}
\label{rhov}
\mathbb{E}_{\rho_k} [ \delta \rho_k(\mu) \delta\rho_k(\nu)] = \mathbb{E}_{\rho_\ell} [ \delta \rho_\ell(\mu)\delta\rho_\ell(\nu) ] = C \rho(\mu) [ \delta_{\mu\nu} - \rho(\nu) ] 
\end{equation}
where $C$ is the certainty measure defined for a general distribution in the main text. Note that there is the same level of certainty in which meaning will be selected as a topic by the signaller as there is in the receiver's beliefs about the interpretation. Correlations between the signaller's and receiver's perspectives are such that
\begin{equation}
\label{rhoc}
\mathbb{E}_{\rho_k,\rho_\ell} [ \delta \rho_k(\mu) \delta\rho_\ell(\nu)] = AC \rho(\mu) [ \delta_{\mu\nu} - \rho(\nu) ]  \;,
\end{equation}
consistent with the definition of the alignment $A$ in the main text. One way to realise such a covariance is for the two meaning distributions to be identical with probability $A$ in any given interaction, and to be sampled independently from a Dirichlet distribution otherwise. However, other constructions in which the two perspectives are always different can give the same covariance behaviour, and our results apply to all of them. 

The strategy now is to expand (\ref{finit}) to second order in $\delta\rho$ and evaluate the averages over the meaning distributions using (\ref{rhov}) and (\ref{rhoc}), thereby generating fitness functions valid for small $C$.  To achieve this, we observe that
\begin{align}
\sum_\mu \rho_{\ell}(\mu) \phi_{\ell} (s|\mu) &= \sum_\mu \rho(\mu) \phi_{\ell} (s|\mu) + \sum_\mu \delta \rho_{\ell}(\mu) \phi_{\ell} (s|\mu) \\
&\approx \phi(s) + \sum_\mu \delta \rho_{\ell}(\mu) \phi_{\ell} (s|\mu) \;,
\end{align}
where we have again used the approximation (\ref{bigMapx}).

Using this expression in (\ref{finit}), and noting that terms of order $\delta\rho$ vanish under the average, we find that, to second order in $\delta\rho$,
\begin{multline}
f_{\ell}(s|m) = 1 + \frac{1}{\rho(m)\phi(s)} \mathbb{E}_{k,\rho_k,\rho_\ell} \left[ \sum_\mu \left( \delta\rho_k(\mu) \phi_k(s|\mu) - \delta\rho_{\ell}(\mu) \phi_\ell(s|\mu) \right) \delta\rho_\ell(m) \right] - \\
\frac{1}{\phi(s)^2} \mathbb{E}_{k,\rho_k,\rho_\ell} \left[\sum_{\mu,\nu} \left( \delta\rho_k(\mu) \phi_k(s|\mu) - \delta\rho_{\ell}(\mu) \phi_\ell(s|\mu) \right) \delta\rho_\ell(\nu) \phi_\ell(s|\nu) \right] \;.
\end{multline}

We can now average over the distributions $\rho_k$ and $\rho_\ell$, making use of  (\ref{rhov}) and (\ref{rhoc}) above. We find
\begin{multline}
f_{\ell}(s|m) = 1+ \frac{C}{\phi(s)} \mathbb{E}_{k} \left[ \sum_\mu (\delta_{m\mu} - \rho(\mu)) (A \phi_k(s|\mu) - \phi_{\ell}(s|\mu) ) \right] - \\
\frac{C}{\phi(s)^2} \mathbb{E}_{k} \left[ \sum_{\mu,\nu} \rho(\mu) (\delta_{\mu\nu} - \rho(\nu)) (A \phi_k(s|\mu) - \phi_{\ell}(s|\mu)) \phi_{\ell}(s|\nu) \right] \;.
\end{multline}
The sums can be performed by using again the approximation (\ref{bigMapx}). This yields
\begin{multline}
f_{\ell}(s|m) = 1+ \frac{C}{\phi(s)} \mathbb{E}_{k} \left[ A \phi_k(s|m) - \phi_{\ell}(s|m) ) \right] +  \\
\frac{C}{\phi(s)^2} \mathbb{E}_{k} \left[ \sum_{\mu} \rho(\mu) (\phi_{\ell}(s|\mu) - A \phi_k(s|\mu)) \phi_{\ell}(s|\mu) \right] \;.
\end{multline}
As before, we can replace $\phi_k(s|m)$ with the society average in the first term, due to the assumption of a well-connected interaction network. Moreover, the central-limit argument that justified the large-$M$ approximation (\ref{bigMapx}) can be applied to the remaining sum over meanings $\mu$. This results in all agent-level quantities in the second term being replaced with society-level averages, removing all dependence on the signaller $k$, and rendering the average over signallers redundant. These computations lead to the expression for the fitness function
\begin{equation}
\label{fab}
f_{\ell}(s|m) =  1+\frac{C}{\phi(s)} \big(  (A \phi(s|m)  + (1-A) \gamma(s))  - \phi_{\ell}(s|m) \big)
\end{equation}
where
\begin{equation}
\gamma(s) = \frac{1}{\phi(s)} \sum_\mu \rho(\mu) \phi(s|\mu)^2 
\end{equation}
which corresponds with the expression given in the main text, Eq.~(\ref{xslf}).

\section{Fitness at the society level}

We now discuss how to generalise the argument that leads to communication being bootstrapped in the absence of feedback to the case of a general number of agents, meanings and signals. This entails determining the fitness $f(s|m)$ of a signalling strategy at the \emph{society} level, taking into account individual differences between agents and how these interact through the nonlinearities in the replicator-mutator equation (\ref{repmut}). To achieve this, we average (\ref{inter}) over the individual agents $\ell$. The key point is that the dependence of the fitness function (\ref{fab}) on the \emph{agent-level} frequency $\phi_{\ell}(s|m)$ is linear. In performing the required average, this means we encounter the quantities $\mathbb{E}_\ell [\phi_{\ell}(s|m)]$,  $\mathbb{E}_\ell [\phi_{\ell}(s|m)\phi_{\ell}(s'|m)]$ and $\mathbb{E}_\ell [\phi_{\ell}(s|m)\phi_{\ell}(s'|m)^2]$. The first of these quantities is the society-level frequency $\phi(s|m)$. The remaining quantities involve the product of agent-level frequencies that are necessarily correlated because their sum over all possible signalling strategies is by definition equal to $1$. We therefore need to model the fluctuations in these local frequencies in a way that takes this constraint into account.

The natural choice, which is expected to be generically valid when the fluctuations are small, is to assume a Dirichlet distribution. Then,
\begin{align}
\label{m2v}
\mathbb{E}_\ell [\phi_{\ell}(s|m)\phi_{\ell}(s'|m)] &= \phi(s|m) [ (1-V) \phi(s'|m) + V \delta_{ss'} ] \\
\mathbb{E}_\ell [\phi_{\ell}(s|m)\phi_{\ell}(s'|m)^2] &= \frac{\phi(s|m)}{1+V} [ (1-V) \phi(s'|m) + V ] [ (1-V) \phi(s'|m) + 2V \delta_{ss'} ]
\end{align}
where the variability $V$ is defined by the covariance matrix
\begin{equation}
{\rm Cov}[\phi_{\ell}(s|m),\phi_{\ell}(s'|m)] = V \phi(s|m) [ \delta_{ss'} - \phi(s'|m)]
\end{equation}
and lies in the range $0\le V \le 1$. This is consistent with the definition of the variability $V$ given in the main text for a general distribution as Eq.~(\ref{V}).  We estimate the value of the variability $V$ that applies for a particular choice of model parameters in the next section below. Given these definitions, the fitness of a signalling strategy in the metapopulation is found from (\ref{fab}) to be
\begin{equation}
\label{fmeta}
f(s|m) = \frac{C}{\phi(s)} \frac{1-V}{1+V} \big[ (1+V) (1-A) \gamma(s)  - V + ((1+V) A - (1-V)) \phi(s|m)  \big] \;,
\end{equation}
This result is equivalent to that previously found by \cite{Cherry2003} when fitness of a species depends linearly on its frequency in the local population.

The significance of this expression is that while the coefficient of the frequency $\phi_{\ell}(s|m)$ in the fitness at the agent level (\ref{fab}) is always negative, at the society level (\ref{fmeta}), the coefficient of $\phi(s|m)$ becomes positive when
\begin{equation}
A > \frac{1-V}{1+V} \;.
\end{equation}
When this equality holds, the strategy that happens to be in the majority will be reinforced, as long as this selective effect can counter the homogenising effects of mutation (which originates in information loss in the agent-based model) and misalignment of perspectives. It is by considering this combination of effects that we arrive at the condition (\ref{thres}) for spontaneous emergence of communication that is reported in the main text.

\section{Estimate of the variability parameter}

The final step in making quantitative predictions for when communication emerges is to estimate the variability $V$. Near the initial non-communicative state, where all signals are equally like to be used for each meaning, we can partition the interactions into those where the interpretation by chance coincides with the topic, and those where it doesn't. When the number of meanings is large, the former event occurs with a probability $p_s = AC + \frac{1-AC}{M}$ and, given that $m$ is the interpretation, the probability that the signal was $\sigma$ is the metapopulation frequency $\phi(\sigma|m)$. When the interpretation differs from the topic (probability $1-p_s$), the conditional probability the signal was $\sigma$ is $\phi(\sigma) \phi_{\ell}(\sigma|m) / \phi_{\ell}(\sigma)$. Approximating $\phi_{\ell}(\sigma)$ as $\phi(\sigma)$, this simplifies to $\phi_{\ell}(\sigma|m)$. Averaging (\ref{update}) over this distribution gives
\begin{align}
\mathbb{E}[\delta \phi_{\ell}(s|m)] &= \frac{\lambda}{1 + \lambda\alpha} p_s [ \phi(s|m) - \phi_{\ell}(s|m) ]  \\
\mathbb{E}[\delta \phi_{\ell}(s|m)\delta \phi_{\ell}(s'|m)] &= \left(\frac{\lambda}{1 + \lambda\alpha}\right)^2  \phi_\ell(s|m) [ \delta_{s,s'} - \phi_{\ell}(s'|m) ] \;.
\end{align}

The stationary distribution of the Kolmogorov equation with these increments is a Dirichlet distribution $P(\{\phi_\ell(s|m)\}) \propto \prod_s \phi_\ell(s|m)^{\gamma-1}$ where the parameter $\gamma= \frac{2p_s(1+\lambda\alpha)}{\lambda}$. Then, for the variability we finally arrive at
\begin{equation}
V=\frac{1}{1+\gamma}= \frac{\lambda}{\lambda + 2p_s(1+\lambda\alpha)}
\end{equation}
with $p_s$ as given above.

Strictly, this estimate is valid only near the non-communicative state. However, we have found reasonable predictions for the location of the communicative fixed point when this variability is assumed to hold more generally, particularly when $C$ is of order $\lambda$. A more complete characterisation of the variability, along with higher-order terms in the fitness function, would help extend the range of validity of our results.

\section{Single-coordinate reduction in a symmetric coordination system}

In the main text, we make use of a simplification of the replicator-mutator equation (\ref{repmut}) when the communication system has a symmetric structure. By this, we mean that, for each meaning $m$, there is one preferred signal $s^\ast(m)$ which is used with frequency $x$, with the remaining frequency $1-x$ distributed evenly among the dispreferred signals. We also take $\rho(m)=\frac{1}{M}$, and work with (\ref{repmut}), averaged over the society, that is, where the local fitness $f_{\ell}(s|m)$ is replaced with the society-level fitness $f(s|m)$ as described above. In the symmetrically-structured state, this fitness $f(s|m)$ takes two different values, $f_+$ if $s=s^*(m)$ and $f_-$ otherwise. The mean fitness $f(m) = x f_+ + (1-x) f_-$. If we now put $s=s^\ast(m)$ into the replicator-mutator equation (\ref{repmut}), we have for its deterministic part that
\begin{equation}
\label{detrepmut}
\dot{x} = \frac{\lambda}{\beta(m) + \lambda\alpha} \frac{1}{M} \left( (f_+ - f_-) x(1-x)  + \lambda \alpha \left[ \frac{1}{S} - x \right] \right) \;.
\end{equation}

For the case of learning with feedback, we have that 
\begin{equation}
f(s|m) = \psi(m|s) = \frac{\phi(s|m)}{\sum_\mu \phi(s|\mu)} = \frac{S}{M} \phi(s|m)
\end{equation}
implying that 
\begin{equation}
f_+ - f_- = \frac{S}{M} \left( x - \frac{1-x}{S-1} \right) = \frac{S^2}{M(S-1)} \left( x - \frac{1}{S} \right) \;.
\end{equation}
Substituting this into (\ref{detrepmut}) we find Eq.~(\ref{xdotf}) in the main text with the parameter $\Gamma=\frac{1}{M}$. For completeness we note that 
\begin{equation}
\label{bm}
\beta(m) = f(m) = x f_+ + (1-x) f_- = \frac{S}{M} \left( x^2 + \frac{(1-x)^2}{S-1} \right)
\end{equation}
although one does not need this to determine the fixed points of the dynamics nor to test their stability.

The case of learning without feedback is a little more involved. We first note that due to the symmetry, the frequency of each signal $s$, $\phi(s)$, is independent of $s$ and is therefore $\phi(s) = \frac{1}{S}$. Moreover, the same symmetry implies that function $\gamma(s)$ that appears in the society-level fitness (\ref{fmeta}) is also independent of $s$. Explicitly, 
\begin{equation}
\gamma(s) = x^2 + \frac{(1-x)^2}{S-1} \;,
\end{equation}
which is  the same as (\ref{bm}), up to a prefactor. Using this in (\ref{fmeta}), we find that
\begin{align}
f_+ - f_- &= C S \frac{1-V}{1+V} [(1+V)A - (1-V)] \left(x - \frac{1-x}{S-1} \right) \\
&= \frac{S^2 C (1-V)}{S-1} \left(A - \frac{1-V}{1+V} \right) \left( x - \frac{1}{S} \right)
\end{align}
which gives us the threshold $\Gamma$ stated as Eq.~(\ref{thres}) in the main text.

\end{document}